%% file: main.tex
\PassOptionsToPackage{table}{xcolor}
\documentclass{article} 
\usepackage{iclr2027_conference,times}
\usepackage{iftex}
\ifXeTeX\usepackage[T1]{fontenc}\fi

\input{math_commands.tex}

\usepackage{hyperref}
\usepackage{url}
\usepackage{enumitem}
\usepackage{booktabs}
\usepackage{graphicx}
\usepackage{amssymb}
\usepackage{multirow}
\usepackage{tabularx}
\usepackage[table]{xcolor}
\newcolumntype{Y}{>{\centering\arraybackslash}X}
\usepackage{wrapfig}
\usepackage{float}
\usepackage{algorithm}
\usepackage{algpseudocode}

\usepackage{textcomp}
\DeclareTextCommandDefault{\textquotedbl}{\char34\relax}
\usepackage{upquote}
\usepackage{tcolorbox}
\tcbuselibrary{breakable,skins,listings}
\definecolor{PromptFrame}{HTML}{3B6E9E}
\definecolor{PromptBack}{HTML}{F4F7FB}
\definecolor{CaseFrame}{HTML}{469C76}
\definecolor{CaseBack}{HTML}{F4FAF7}
\definecolor{CaseLabel}{HTML}{30485C}
\newtcblisting{promptbox}[1]{%
  breakable, enhanced, listing only,
  colback=PromptBack, colframe=PromptFrame, coltitle=white,
  fonttitle=\small\bfseries, title={#1},
  boxrule=0.6pt, arc=1.5pt, left=4pt, right=4pt, top=3pt, bottom=3pt,
  listing options={basicstyle=\ttfamily\scriptsize, breaklines=true, breakindent=0pt,
    breakautoindent=false, postbreak={\mbox{\textcolor{PromptFrame}{$\hookrightarrow$}\space}},
    columns=fullflexible, keepspaces=true, showstringspaces=false,
    upquote=true, aboveskip=0pt, belowskip=0pt}}
\newtcolorbox{casebox}[1]{%
  breakable, enhanced, colback=CaseBack, colframe=CaseFrame, coltitle=white,
  fonttitle=\small\bfseries, title={#1},
  boxrule=0.6pt, arc=1.5pt, left=5pt, right=5pt, top=4pt, bottom=4pt}
\newtcblisting{caselisting}[1]{%
  breakable, enhanced, listing only,
  colback=CaseBack, colframe=CaseFrame, coltitle=white,
  fonttitle=\small\bfseries, title={#1},
  boxrule=0.6pt, arc=1.5pt, left=4pt, right=4pt, top=3pt, bottom=3pt,
  listing options={basicstyle=\ttfamily\scriptsize, breaklines=true, breakindent=0pt,
    breakautoindent=false, postbreak={\mbox{\textcolor{CaseFrame}{$\hookrightarrow$}\space}},
    columns=fixed, basewidth=0.5em, keepspaces=true, showstringspaces=false,
    upquote=true, aboveskip=0pt, belowskip=0pt}}
\newcommand{\caseturn}[2]{%
  \par\smallskip\noindent{\footnotesize\textcolor{CaseLabel}{\textsf{\textbf{#1}}}\enspace #2}\par}

\input{revision_macros}

\definecolor{gaincolor}{RGB}{34,139,34}
\definecolor{dropcolor}{RGB}{220,20,60}
\newcommand{\gp}[1]{\raisebox{-0.5ex}{\textcolor{gaincolor}{\tiny$\uparrow$#1}}}
\newcommand{\gn}[1]{\raisebox{-0.5ex}{\textcolor{dropcolor}{\tiny$\downarrow$#1}}}

\title{Travel Back To Move Forward: Retrospective Exploration and Evolving for Long Horizon Agents}

\author{%
Ziang Ye$^{1,2,6}$\thanks{Work done during internship at Meituan.},\enspace Wentao Shi$^{3}$,\enspace Lei Chen$^{1}$,\enspace
Zhengzhou Cai$^{1}$,\enspace Yu Wang$^{1}$, \\
\bfseries Zhiyuan Yao$^{4}$,\enspace Zihan Wang$^{5}$,\enspace Xierui Song$^{6}$\thanks{Corresponding authors},\enspace
Fuli Feng$^{1}$,\enspace Yixin Cao$^{7}$\footnotemark[2] \\[0.5em]
\normalfont $^{1}$University of Science and Technology of China\enspace
$^{2}$Shanghai Innovation Institute \\
\normalfont $^{3}$Facebook\enspace $^{4}$Zhejiang University\enspace $^{5}$Tsinghua University\enspace
$^{6}$Meituan\enspace $^{7}$Fudan University \\[0.4em]
\normalfont \texttt{yza03@mail.ustc.edu.cn},\enspace \texttt{guqi03@meituan.com}, \\
\normalfont \texttt{fulifeng93@gmail.com},\enspace \texttt{caoyixin2011@gmail.com}
}

\iclrfinalcopy
\newlength{\headlogoband}
\newcommand{\headlogo}[2]{\raisebox{\dimexpr(\headlogoband-#1)/2\relax}{\includegraphics[height=#1]{#2}}}
\newcommand{\letterhead}{%
  \noindent\makebox[\textwidth][s]{%
    \headlogo{0.9cm}{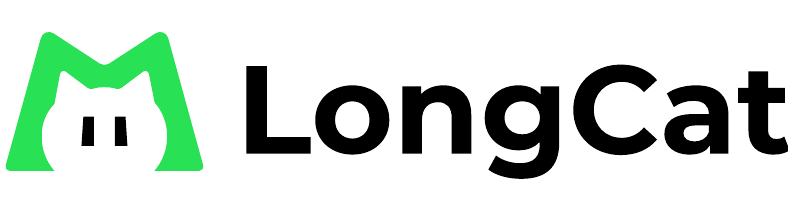}\hfill
    \headlogo{0.9cm}{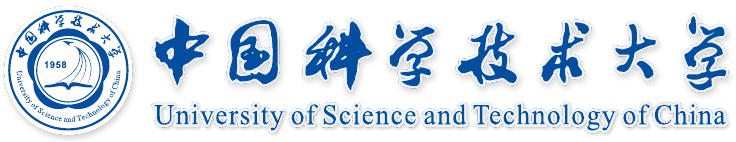}\hspace{2.2em}%
    \headlogo{0.86cm}{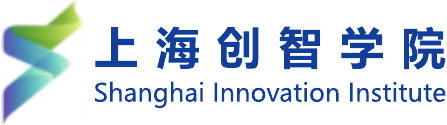}}\par
  \nointerlineskip\vspace{4pt}%
  \noindent\rule{\textwidth}{0.4pt}\par}
\fancypagestyle{preprintfirst}{%
  \fancyhead{}%
  \renewcommand{\headrulewidth}{0pt}}
\makeatletter
\let\iclr@orig@maketitle\maketitle
\renewcommand{\maketitle}{%
  \vspace*{-2.4cm}\letterhead\vspace{0.45cm}%
  \iclr@orig@maketitle\lhead{Preprint}\thispagestyle{preprintfirst}}
\makeatother
\begin{document}

\maketitle
\input{sections/abstract}

\input{sections/introduction}

\input{sections/related_work}

\input{sections/preliminary}

\input{sections/method}

\input{sections/experiments}

\input{sections/analysis}

\input{sections/conclusion}

\label{review:main_end}
\input{sections/ai_statement}
\input{sections/ethics_statement}
\input{sections/reproducibility_statement}

\nocite{*}
\bibliography{iclr2026_conference}
\bibliographystyle{iclr2027_conference}

\appendix
\input{sections/appendix}

\end{document}

%% file: math_commands.tex
\usepackage{amsmath,amsfonts,bm}

\def\eqref#1{equation~\ref{#1}}

\def\1{\bm{1}}

\DeclareMathAlphabet{\mathsfit}{\encodingdefault}{\sfdefault}{m}{sl}
\SetMathAlphabet{\mathsfit}{bold}{\encodingdefault}{\sfdefault}{bx}{n}



%% file: revision_macros.tex
\providecommand{\RevisionMode}{0}
\usepackage{xparse}
\definecolor{RevisionOld}{RGB}{125,125,125}
\definecolor{RevisionNew}{RGB}{26,75,174}
\NewDocumentCommand{\Rev}{m +m +m}{%
  \ifnum\RevisionMode=1\relax
    \begingroup\color{RevisionOld}%
    \renewcommand{\label}[1]{}
    \renewenvironment{equation}{\begin{equation*}}{\end{equation*}}
    #2\par\endgroup
    \begingroup\color{RevisionNew}\noindent{\scriptsize\textsf{[#1]}}\enspace#3\par\endgroup
  \else\ifnum\RevisionMode=2\relax#2\else#3\fi\fi}
\newcommand{\RevisionPlainGains}{%
  \renewcommand{\gp}[1]{\raisebox{-0.5ex}{\tiny$\uparrow$##1}}%
  \renewcommand{\gn}[1]{\raisebox{-0.5ex}{\tiny$\downarrow$##1}}}
\NewDocumentCommand{\RevInline}{m m m}{%
  \ifnum\RevisionMode=1\relax{\color{RevisionOld}\RevisionPlainGains#2}\enspace{\color{RevisionNew}\RevisionPlainGains#3}%
  \else\ifnum\RevisionMode=2\relax#2\else#3\fi\fi}
\NewDocumentCommand{\RevAdd}{m +m}{%
  \ifnum\RevisionMode=2\relax\else
    \ifnum\RevisionMode=1\relax\begingroup\color{RevisionNew}\noindent{\scriptsize\textsf{[#1]}}\enspace#2\par\endgroup
    \else#2\fi\fi}
\NewDocumentCommand{\AuthorQuery}{m +m}{%
  \ifnum\RevisionMode=1\relax\par\smallskip
    {\color{RevisionNew}\small\noindent\textbf{Author input [#1].} #2\par}\smallskip
  \fi}
\NewDocumentCommand{\RevTitle}{m +m +m}{%
  \ifnum\RevisionMode=1\relax{\color{RevisionOld}#2}\par\vspace{0.35em}{\color{RevisionNew}#3}%
  \else\ifnum\RevisionMode=2\relax#2\else#3\fi\fi}
\NewDocumentCommand{\RevNext}{m +m +m +m}{%
  \ifnum\RevisionMode=2\relax#2\else
    \Rev{#1}{\textsf{\scriptsize[Original]}\enspace#2\par
      \textsf{\scriptsize[Previous proposal]}\enspace#3}{#4}\fi}
\NewDocumentCommand{\RevInlineNext}{m +m +m +m}{%
  \ifnum\RevisionMode=2\relax#2\else
    \RevInline{#1}{#2\enspace/\enspace#3}{#4}\fi}
\NewDocumentCommand{\RevTitleNext}{m +m +m +m}{%
  \ifnum\RevisionMode=2\relax#2\else
    \RevTitle{#1}{#2\par#3}{#4}\fi}

%% file: sections/abstract.tex
\begin{abstract}

Large language model agents increasingly operate in long-horizon environments where early mistakes can trigger cascading failures and trap agents in dead-end trajectories. Agents must therefore be able to proactively recover from such states and turn their exploratory experience into guidance for subsequent decisions. We introduce \textbf{ReTRAIL}, a framework that enables agents to rewind the environment to earlier states while preserving valid progress and carrying forward knowledge learned from abandoned exploration. The environment state moves backward while the agent's knowledge continues to move forward. ReTRAIL trains a shared policy with two complementary signals: State-Anchored GRPO compares alternative continuations from revisited states to improve execution, while a branch-advancement reward guides reflection through the progress of subsequent exploration. Experiments across four interactive benchmark MineSweeper, Sokoban, ALFWorld and ScienceWorld show that simply adding a rewind primitive provides limited benefit and can even degrade performance. In contrast, ReTRAIL enables agents to rewind from unproductive states and use insights from abandoned branches to guide subsequent exploration. These results demonstrate the value of coupling environment rewinding with experience reuse within an ongoing task for long-horizon task solving. The code is available at \url{https://github.com/whi497/ReTRAIL}.
\end{abstract}

%% file: sections/introduction.tex
\section{Introduction}
Large language model (LLM) agents are rapidly evolving from conversational systems into autonomous decision-making agents capable of solving long-horizon tasks through extended interactions, diverse tool use, and complex state transitions~\citep{swebench,sun2026agentsexam,vitabench}. As task complexity increases, agents inevitably encounter unfamiliar constraints and make early mistakes that may alter the environment, trigger cascading failures, and accumulate irrelevant information into the context~\citep{zhou2024lats,ye2026lookleapautonomousexploration}. Recovering from such failures requires agents to revisit earlier viable environment states without discarding either valid progress or useful experience. Crucially, returning to an earlier state should not mean returning to the same decisions: agents must distill what they have learned from abandoned exploration and use it to guide a more informed continuation.

Prior work has sought to guide LLMs to recover from errors and learn from unsuccessful history through reflection, repeated trials, and backtracking. Interleaving reasoning with acting lets an agent revise its plan at each step, but the revision stays local and the execution keeps moving forward~\citep{react_paradigm,choi2026reactree}. Cross-episode adaptation uses accumulated experience to guide new attempts~\citep{shinn2023reflexion,ouyang2026reasoningbank,jiang2026lamer}, but applies it after an episode terminates and execution restarts from the initial state, requiring earlier progress to be repeated. Backtracking and trajectory-revision methods enable agents to reconsider earlier decisions through tree search, prompted controllers, or supervised correction traces~\citep{yuan2025agentr,li2025generatorassistantstepwiserollbackframework,zhang2026webrollback}. A central challenge is to train agents to proactively recover within an ongoing task and use the experience gained from abandoned branches to improve subsequent exploration. This calls for learning signals that improve both execution after a rewind and the experience carried forward.

Our key observation is that environment states can be rewound while the knowledge gained from exploration should be accumulated. An unsuccessful branch may reveal hidden constraints, ineffective actions, or promising alternatives, allowing the agent to revisit an earlier state with a better-informed context. We introduce \textbf{ReTRAIL}, a framework that exposes an environment-level \textit{rewind} interface to the agent, allowing it to restore earlier environment states. Building on this interface, we propose \textbf{Retrospective Exploration}: the agent proactively leaves an unproductive branch, rewinds the environment to an earlier state, distills useful information from the abandoned branch into a compact reflection, and resumes execution with the preserved execution prefix and accumulated experience. To jointly develop the agent’s ability to rewind and distill useful experience from abandoned branches, we use \textbf{State-Anchored GRPO} to improve execution by comparing alternative continuations from the same environment state, and a \textbf{branch-advancement reward} to train reflection based on the relative progress of subsequent exploration.

We evaluate ReTRAIL across four diverse interactive benchmarks: MineSweeper~\citep{Li_2024}, Sokoban~\citep{racaniere2017sokoban}, ALFWorld~\citep{shridhar2021alfworld}, and ScienceWorld~\citep{wang2022scienceworld}. ReTRAIL achieves the highest success rates among the evaluated methods when rewind is enabled. Our experiments reveal that simply providing existing agents with a rewind operation does not consistently improve performance and can even degrade it. In contrast, ReTRAIL benefits from proactive rewinding across all four benchmarks and achieves further gains through reflection-guided retries. On MineSweeper, success increases from 58.8\% without rewind to 63.5\% with proactive rewinding, and reaches 95.3\% with up to three reflection-guided attempts, compared with 70.8\% for the strongest baseline under the same evaluation protocol. Further analysis shows that retry gains are larger on longer tasks and when abandoned branches reveal information useful for subsequent decisions. These findings highlight the value of learning to rewind and carry forward useful experience, enabling unsuccessful exploration to inform continued progress within an ongoing long-horizon task.

Our main contributions are summarized as follows:
\begin{itemize}[leftmargin=*, parsep=1pt]
    
    \item We introduce \textbf{ReTRAIL}, a framework for \textbf{Retrospective Exploration} through an environment-level rewind interface. Agents proactively revisit earlier states while preserving valid progress and carrying forward experience distilled from abandoned branches to guide subsequent exploration.
    \item We develop two complementary learning signals for a shared execution and reflection policy. \textbf{State-Anchored GRPO} compares alternative continuations from the same environment state to improve execution, while a \textbf{branch-advancement reward} trains reflection based on the relative progress of subsequent exploration.
    \item We evaluate ReTRAIL across four interactive benchmarks, demonstrating improved success through proactive rewinding and reflection-guided retries. Further experiments show larger gains on longer and harder tasks, highlighting the value of retaining experience to guide subsequent exploration.
\end{itemize}

%% file: sections/related_work.tex
\section{Related Work}

\subsection{Long-Horizon Execution and Recovery}

Sustaining reliable execution over long interaction sequences motivates research on planning, context management, and error recovery in LLM agents~\citep{react_paradigm,autocot,yuan2025agentr}.
Search-based methods explore alternative action sequences within context~\citep{zhou2024lats, choi2026reactree}.
Agent-R learns decision correction from revision traces~\citep{yuan2025agentr}.
However, revising decisions through context alone does not reverse changes to the environment, which can leave agents trapped in unproductive states. WebRollback uses action, critique, and rollback modules to select recovery points and carry forward experience summaries, with prompting and supervised fine-tuning from demonstrations~\citep{zhang2026webrollback}.
GA-Rollback uses a prompted assistant to detect and localize errors during inference, restores earlier environment states through reset-and-replay, and retains error feedback to guide subsequent actions~\citep{li2025generatorassistantstepwiserollbackframework}.
ReTRAIL take a step forward to train a shared execution and reflection policy that couples environment-level rewind with reinforcement learning, enabling more diverse exploration and more effective interaction budgets using.

\subsection{Test-Time Adaptation and Self-Evolving Agents}

Adapting from interaction experience enables LLM agents to improve their behavior in unfamiliar tasks and environments.
Reflection and memory mechanisms distill past trajectories into feedback and reusable strategies that guide subsequent decisions~\citep{shinn2023reflexion,ouyang2026reasoningbank}.
Beyond memory accumulation, EvoTest and Meta-TTL adapt prompts, memory, and tool-use routines from episode feedback ~\citep{he2026evotest,lou2026metattl}.
Recent work further uses reinforcement learning to optimize how agents explore and adapt from experience.
LaMer and MR-Search train exploration and post-episode reflection through meta-reinforcement learning, using discounted outcome reward to provide feedback for adaptation~\citep{jiang2026lamer,xiao2026mrsearch}.
Learning to Self-Evolve rewards context revisions according to downstream performance improvements~\citep{chen2026lse}.
ReTRAIL brings experience-informed adaptation into recovery within an ongoing task.
By revisiting earlier environment states while preserving valid progress and distilled branch experience, agents can revise their exploration before the current attempt terminates, with subsequent branch progress providing a learning signal for reflection.

%% file: sections/preliminary.tex
\section{Retrospective Exploration}
\label{sec:autonomous-execution}

We formalize \textbf{Retrospective Exploration}, in which an agent
rewinds to an earlier environment state while preserving the
execution prefix and retaining experience from abandoned branches.
We first introduce agent interaction and multi-episode adaptation,
then describe how an environment-level \textit{rewind} interface
enables experience-informed recovery within an ongoing task.

\begin{figure}[t]
    \centering
    \includegraphics[width=0.93\linewidth]{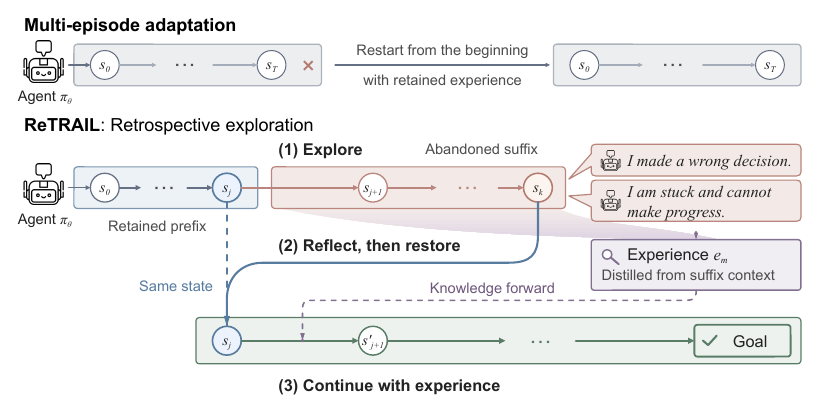}
    \caption{\textbf{Retrospective exploration with retained experience.}
Multi-episode adaptation (top) restarts the task from \(s_0\) after an unsuccessful attempt, carrying retained experience into a new episode.
In contrast, ReTRAIL (bottom) proactively abandons an unproductive branch, distills its experience, rewinds to an earlier state, and resumes with the retained execution prefix and accumulated experience.
Recovery can occur before terminal failure.}
    \label{fig:flow}
    \vspace{-10pt}
\end{figure}

\paragraph{Agent Interaction and Multi-Episode Adaptation.}
Given a task \(u\), an LLM agent starts from environment state \(s_0\).
At step \(t\), it observes \(o_t\) and selects an action
\(a_t\in\mathcal A\) based on the interaction history
\(h_t=(o_0,a_0,\ldots,a_{t-1},o_t)\).
The environment transitions to \(s_{t+1}\), returning reward
\(r_t\) and observation \(o_{t+1}\).
An episode yields a trajectory
\(\tau=(s_0,a_0,r_0,\ldots,a_{T-1},r_{T-1},s_T)\),
ending upon task completion, environment termination, or
exhaustion of its interaction budget.

Multi-episode adaptation groups successive attempts into a trial
\(\mathcal T=(\tau^{(0)},\ldots,\tau^{(N-1)})\)~\citep{jiang2026lamer}.
Each episode restarts from \(s_0\), while the policy conditions on
experience \(\mathcal E^{(n)}\) retained from preceding episodes:
\begin{equation}
    a_t^{(n)}
    \sim \pi_\theta
    \bigl(\cdot\mid u,h_t^{(n)},\mathcal E^{(n)}\bigr),
    \qquad \mathcal E^{(0)}=\varnothing.
    \label{eq:multi_episode_policy}
\end{equation}
The trial ends upon success or exhaustion of the attempt budget.
Experience thus accumulates across episodes, but each restart
requires rebuilding previously achieved environment progress.

\paragraph{Retrospective Exploration.}
ReTRAIL provides an environment-level \textit{rewind} interface that allows the agent to restore a previously visited state during an ongoing attempt. When the current branch is unlikely to yield further progress, the agent can proactively rewind, preserve the execution prefix, and use experience distilled from the abandoned branch to guide a new continuation. We call this process \emph{retrospective exploration}.

\textbf{Execute.} A forward segment \(\sigma_m\) is the contiguous interaction sequence from the initial state or a restored state to the next recovery boundary or trial termination. Rewinding ends the current branch and starts a new segment from the selected earlier state, while accumulated experience remains available. We index these segments by \(m\) and condition execution on the interaction history \(h_t\) and accumulated branch experience
\(\mathcal E_m=(e_0,\ldots,e_{m-1})\), where \(e_i\) is the reflection distilled at recovery \(i\) and \(\mathcal E_0=\varnothing\):
\begin{equation}
    a_t\sim\pi_\theta(\cdot\mid u,h_t,\mathcal E_m).
    \label{eq:experience_conditioned_execution}
\end{equation}
\textbf{Detect recovery.} At \(s_{k_m}\), recovery may be initiated proactively once the agent judges that the current branch is unlikely to yield further progress, or triggered by a failure signal.
The agent selects an earlier state \(s_{j_m}\), partitioning the current execution path into a retained prefix and an abandoned suffix:
\begin{equation}
\begin{aligned}
    \rho_m
    &=
    \operatorname{rewind}
    \left(s_{k_m}\rightarrow s_{j_m}\right),
    \qquad j_m<k_m,\\
    \tau_{0:k_m}
    &=
    \underbrace{\tau_{0:j_m}}_{\text{retained prefix}}
    \oplus
    \underbrace{\beta_m}_{\text{abandoned suffix}},
    \qquad
    \beta_m=\tau_{j_m:k_m}.
\end{aligned}
\label{eq:rewind_partition}
\end{equation}
Here, \(\rho_m\) denotes the rewind decision, before restoration is executed.
The indices \(j_m\) and \(k_m\) refer to positions on the current execution path.
The operator \(\oplus\) denotes concatenation, with a shared boundary state included only once when joining trajectory segments.

\textbf{Reflect.} Before discarding \(\beta_m\), the agent distills its useful information into a compact reflection:
\begin{equation}
    e_m
    \sim
    \pi_\theta
    \left(
        \cdot
        \mid
        u,h_{j_m},\beta_m,\mathcal E_m
    \right).
    \label{eq:branch_experience}
\end{equation}

The reflection may capture discovered constraints, failure causes, actions to avoid, or alternative directions.

\textbf{Restore and resume.} Return to \(s_{j_m}\), truncate history to \(h_{j_m}\), and retain experience:
\par\nopagebreak[4]
\noindent
\begin{minipage}{0.32\linewidth}
\begin{equation}
    \mathcal E_{m+1}=\mathcal E_m\oplus e_m.
\end{equation}
\end{minipage}\hfill
\begin{minipage}{0.65\linewidth}
\begin{equation}
    \big(s_{k_m},h_{k_m},\mathcal E_m\big)
    \;\longrightarrow\;
    \big(s_{j_m},h_{j_m},\mathcal E_{m+1}\big).
    \label{eq:rewind_resume}
\end{equation}
\end{minipage}
\par\nopagebreak[4]
Execution resumes with the retained prefix and updated experience, allowing subsequent decisions to use a better-informed context from the same environment state.
A complete trial of retrospective exploration is written as
\begin{equation}
    \Gamma
    =
    \big(
        \sigma_0,\rho_0,e_0,
        \sigma_1,\rho_1,e_1,
        \ldots,
        \sigma_M
    \big),
    \label{eq:rollback_trial}
\end{equation}
where \(\sigma_m\) denotes a forward trajectory segment ending at a recovery boundary or trial termination.
The first segment starts from \(s_0\); each subsequent segment resumes from the selected earlier state and conditions on the updated experience.
The abandoned suffix \(\beta_m\) is determined by the selected rewind point and may span portions of multiple earlier segments.

%% file: sections/method.tex
\section{Learning from Retrospective Exploration}
\label{sec:policy_optimization}

Retrospective exploration allows agents to revisit earlier states and continue with accumulated experience.
Learning to use this capability effectively requires improving both execution decisions and the experience distilled from abandoned branches.
It therefore yields two complementary learning signals.
First, alternative continuations from the same environment state and retained prefix can be compared to identify more productive execution paths.
Second, the progress of a resumed continuation can be compared with that of the abandoned suffix to assess the usefulness of the experience carried forward.

\begin{figure}[t]
    \centering
    \includegraphics[width=0.95\linewidth]{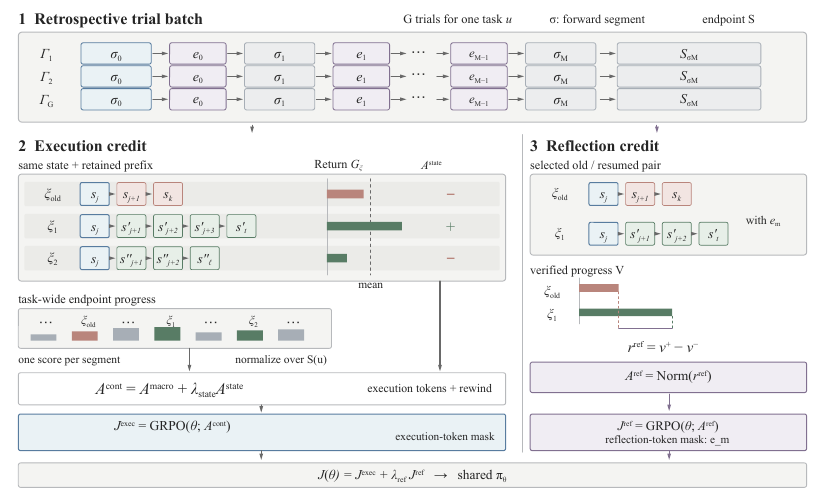}
    \caption{\textbf{Learning from retrospective exploration.}
    The top panel collects retrospective trials for one task.
    Left: State-Anchored GRPO combines task-wide endpoint progress with return comparisons among continuations sharing an environment state and retained prefix, producing \(A^{\mathrm{cont}}\) for execution tokens including rewind.
    Right: the verified-progress difference between a matched abandoned/resumed pair produces \(A^{\mathrm{ref}}\) for the reflection tokens that generate \(e_m\).
    The separate masked objectives jointly update one shared policy. }
    \label{fig:rollback-optimization}
    \vspace{-6pt}
\end{figure}

We use these two signals to jointly train a shared execution and reflection policy.
\textbf{State-Anchored GRPO} implements the first comparison by assigning credit to continuations from the same state and retained prefix.
A \textbf{branch-advancement reward} implements the second by measuring the progress of the resumed continuation relative to the abandoned suffix.
Figure~\ref{fig:rollback-optimization} illustrates these two learning signals.

\subsection{State-Anchored GRPO over Revisited States}
\label{sec:state_anchored_grpo}

A final task outcome gives limited guidance about which decisions were productive. Local comparisons, as used in GiGPO~\citep{feng2025gigpo}, provide more focused feedback. Retrospective exploration creates the comparison groups needed for execution credit: after returning to an earlier point, the agent can generate alternative continuations from the same environment state while retaining the same execution prefix.

\paragraph{Constructing continuation groups.}
For each task \(u\), we sample a group of retrospective trials
\(\{\Gamma_i\}_{i=1}^{B}\) and collect their forward segments in \(\mathcal S(u)\).
Each recorded state visit defines a continuation \(\xi\) ending with its segment at success, failure, rewind, or budget exhaustion.
A model-initiated rewind is included as the terminal execution decision when present.

Let \(s_\xi\) and \(h_\xi\) denote the starting environment state and retained history of \(\xi\).
We group continuations by the same task, starting state, and retained history:
\begin{equation}
    \mathcal H_u(s,h)
    =
    \left\{
        \xi\in\mathcal C(u):
        s_\xi=s,\ h_\xi=h
    \right\},
    \label{eq:continuation_group}
\end{equation}
where \(\mathcal C(u)\) is the collection of continuations extracted from the sampled trials.
Groups include earlier segment suffixes passing through \(s\) and new continuations after rewind.
Samples retain their original generation history and experience; the accumulated experience may differ within a group.

\paragraph{Assigning execution credit.}
Each continuation has a discounted return \(G_\xi\) from its starting state to the segment endpoint and a task-progress score \(S_\xi\) at that endpoint, shared by continuations from the same segment.
Its advantage combines task-wide and state-relative feedback:
\begin{equation}
    A_\xi^{\mathrm{cont}}
    =
    \underbrace{
        \operatorname{Norm}_{\mathcal S(u)}(S_\xi)
    }_{A_\xi^{\mathrm{macro}}}
    +
    \lambda_{\mathrm{state}}
    \underbrace{
        \operatorname{Norm}_{\mathcal H_u(s_\xi,h_\xi)}(G_\xi)
    }_{A_\xi^{\mathrm{state}}},
    \label{eq:state_anchored_advantage}
\end{equation}
where \(\operatorname{Norm}\) subtracts the corresponding group mean and divides by its standard deviation plus \(\epsilon\).
The macro term normalizes endpoint progress over \(\mathcal S(u)\), counting each segment once, to preserve the task-wide objective.
The state-relative term favors higher-return continuations from the same state and prefix.
For singleton groups, this term is zero and the macro signal remains.

We apply \(A_\xi^{\mathrm{cont}}\) to the execution tokens in \(\xi\), including any terminal model-initiated rewind decision.
The retained prefix serves as input context and is excluded from the continuation loss.
Execution returns end at rewind; subsequent advancement supervises the reflection response. Appendix~\ref{app:state_anchored_grpo} details the construction, and Appendix~\ref{app:credit_example} illustrates this credit boundary.

\subsection{Learning Reflection from Branch Advancement}
\label{sec:reflection_learning}

Reflection is generated from an abandoned branch, but its value is realized only through the continuation that follows recovery. The discarded suffix may contain useful information about invalid actions, hidden constraints, or promising alternatives; however, the final trial outcome is too coarse and too delayed to determine whether this information improved the resumed exploration. A successful trial may succeed despite an uninformative reflection, while a useful reflection may produce measurable progress before task completion. We therefore evaluate reflection by comparing the progress of the post-recovery continuation with that of the abandoned branch from the same restored state.

At recovery \(m\), the agent distills the abandoned suffix \(\beta_m\) into experience \(e_m\) and resumes from state \(s_{j_m}\).
The abandoned suffix and the new forward segment \(\sigma_{m+1}\) share this starting state, while the latter conditions on the updated experience \(\mathcal E_{m+1}\).
Here, \(\beta_m\) is the suffix discarded by the rewind and may contain portions of multiple earlier segments; it need not coincide with a single continuation used for execution grouping.

Using an environment-verifiable progress measure \(V\), we evaluate the two paths:
\begin{equation}
    v^-_m
    =
    V\!\left(\beta_m\mid s_{j_m}\right),
    \qquad
    v^+_m
    =
    V\!\left(\sigma_{m+1}\mid s_{j_m}\right),
    \label{eq:segment_progress}
\end{equation}
where \(V\) measures task progress, such as completed subgoals, for exploration starting from \(s_{j_m}\).
The branch-advancement reward uses their difference as a proxy for experience utility; positive values indicate greater progress after recovery:
\begin{equation}
    r_m^{\mathrm{ref}}
    =
    v^+_m-v^-_m.
    \label{eq:reflection_reward}
\end{equation}
\begin{samepage}
\paragraph{Joint policy optimization.}
We jointly optimize execution and reflection through
\begin{equation}
    \mathcal J(\theta)
    =
    \mathcal J_{\mathrm{GRPO}}^{\mathrm{exec}}
    \left(\theta;A^{\mathrm{cont}}\right)
    +
    \lambda_{\mathrm{ref}}
    \mathcal J_{\mathrm{GRPO}}^{\mathrm{ref}}
    \left(\theta;A^{\mathrm{ref}}\right),
    \label{eq:retrail_objective}
\end{equation}
where \(A^{\mathrm{ref}}\) is obtained by group-normalizing the branch-advancement rewards \(r_m^{\mathrm{ref}}\), and
\(\lambda_{\mathrm{ref}}\) controls the relative weight of reflection training.
The two objectives update \(\pi_\theta\) through execution tokens (including model-initiated rewind) and reflection tokens, respectively.
Appendix~\ref{app:state_anchored_grpo} gives sample weighting and GRPO details.
\par
\end{samepage}

%% file: sections/experiments.tex
\section{Experiments}
\label{sec:experiments}

Our experiments ask whether ReTRAIL improves recovery-enabled success while retaining competitive forward execution (Section~\ref{sec:main-results}), which rollout and learning components support these gains (Section~\ref{sec:analysis}), and how performance varies with instance difficulty, task length, and the number of recovery attempts (Sections~\ref{sec:scaling} and~\ref{sec:stratified}).

\subsection{Experimental Setup}
\label{sec:exp-setup}

\paragraph{Environments.}
The reported results cover four interactive benchmarks with complementary failure structures. \textbf{MineSweeper}~\citep{Li_2024} requires logical deduction under partial observability; unsuccessful exploration may reveal board constraints. \textbf{Sokoban}~\citep{racaniere2017sokoban} tests planning with irreversible forward transitions, such as pushing a box into a dead corner. \textbf{ALFWorld}~\citep{shridhar2021alfworld} evaluates multi-step instruction following, while \textbf{ScienceWorld}~\citep{wang2022scienceworld} involves procedural reasoning with long dependency chains. All use text interaction and the state-restoration interface in Section~\ref{sec:autonomous-execution}. Dataset partitions are listed in Table~\ref{tab:dataset_partitions}.

\paragraph{Baselines.}
We compare with zero-shot prompting, ReAct~\citep{react_paradigm}, Reflexion~\citep{shinn2023reflexion}, adapted GA-Rollback~\citep{li2025generatorassistantstepwiserollbackframework}, and trained baselines GRPO~\citep{GRPO}, RLOO~\citep{ahmadian2024basicsrevisitingreinforcestyle}, GiGPO~\citep{feng2025gigpo}, BEACON~\citep{wang2026beacon}, and LaMer~\citep{jiang2026lamer}. GiGPO uses local group comparisons; LaMer learns across episodes. Recovery-enabled evaluations share a rewind interface with corresponding execution and reflection prompts. 

\paragraph{Evaluation protocol and metrics.}
\textbf{Pass@1 (P@1)}  measures single-pass execution without rewind. \textbf{Trial@1 (T@1)} permits model-initiated rewind but terminates at the first irreversible failure; one macro-attempt may therefore explore multiple branches. \(\Delta_{\mathrm{pro}}=\mathrm{T@1}-\mathrm{P@1}\) measures the net gain from enabling proactive recovery. \textbf{Trial@3 (T@3)} permits up to three macro-attempts, with reflection-guided recovery after irreversible failure. We denote \(\Delta_{\mathrm{ref}}=\mathrm{T@3}-\mathrm{T@1}\) as the \emph{retry gain}: it measures the combined benefit of additional attempts, recovery, and retained experience. The configured step limit for all settings is 60. Details are provided in Appendix~\ref{app:control_flow}.

\paragraph{Training details.}
We use Qwen3-4B~\citep{yang2025qwen3} as the base model for all methods.
ReTRAIL samples groups of trials using retrospective exploration and is optimized using our State-Anchored GRPO
objective (Section~\ref{sec:state_anchored_grpo}).
For the standard RL baselines, we adjust the rollout group size to match the sampling budget of
ReTRAIL at each gradient update.
We otherwise keep the learning rate, batch size, prompt template, and task splits identical across
methods. The continuation construction and execution objective are detailed in Appendix~\ref{app:state_anchored_grpo}. Appendix~\ref{app:seed_aggregation} reports equal-benchmark aggregation.
Reported results are means over three evaluation seeds.

\subsection{Main Results}
\label{sec:main-results}

\input{tables/main_results}

\paragraph{ReTRAIL achieves stronger performance.}
ReTRAIL achieves the highest T@1 and T@3 on all four benchmarks
(Table~\ref{tab:main_results}), outperforming the strongest baseline
in T@3 by \(24.5\) percentage points on MineSweeper and \(10.2\)
on ScienceWorld. Its forward-only P@1 remains competitive,
slightly exceeding the strongest baseline on MineSweeper, ALFWorld,
and ScienceWorld, while trailing GiGPO by \(2.3\) points on Sokoban.
Together, these results show that ReTRAIL's advantage becomes more
pronounced when recovery is enabled, while its forward-solving
ability remains competitive.

\paragraph{ReTRAIL makes effective use of test-time recovery.}
ReTRAIL benefits from recovery both within and across macro-attempts.
Within a single attempt, proactive recovery yields positive gains
on every benchmark, ranging from \(1.3\) percentage points on
ALFWorld to \(10.9\) on Sokoban, whereas enabling rewind for baseline
policies produces mixed changes. Across attempts, ReTRAIL further
improves with reflection-guided recovery and retained experience.
On MineSweeper, success rises from \(63.5\) at T@1 to \(95.3\)
at T@3, a gain of \(31.8\) points, compared with \(9.1\) for GiGPO
and \(13.6\) for LaMer. These results demonstrate ReTRAIL's ability
to adapt through recovery, benefiting from both within-attempt
course correction and continued exploration after failure.

\begin{figure}[H]
  \centering
  \providecommand{\explorationefficiencyfigure}{imgs/exploration_efficiency.pdf}
  \includegraphics[width=0.75\textwidth]{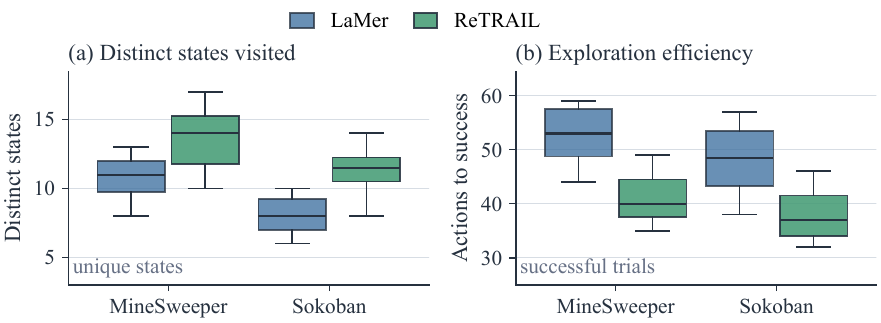}
  \caption{Exploration diversity and efficiency of LaMer and ReTRAIL on MineSweeper and Sokoban.
  (a) States visited, where higher values indicate broader exploration;
  (b) Actions to success, where lower values indicate more efficient
  solving. ReTRAIL visits more states while requiring fewer actions.}
  \label{fig:exploration_efficiency}
  \vspace{-8pt}
\end{figure}

\paragraph{ReTRAIL improves exploration coverage and execution efficiency.}
Figure~\ref{fig:exploration_efficiency} compares the coverage of distinct
environment states and the number of actions taken in successful trials.
Relative to LaMer, ReTRAIL increases the median number of distinct
states visited from \(11.0\) to \(14.0\) on MineSweeper and from
\(8.0\) to \(11.5\) on Sokoban. Meanwhile, the median action count
among successful trials decreases from \(53.0\) to \(40.0\) and
from \(48.5\) to \(37.0\), respectively. Together, these findings
suggest that recovering early from unproductive branches helps
ReTRAIL explore more broadly while completing tasks more efficiently.

\subsection{Scaling Instance Difficulty and Recovery Attempts}
\label{sec:scaling}

We next examine performance as instance difficulty and inference-time computation increase.

\begin{figure*}[t]
  \centering
  \providecommand{\scalingrecoveryfigure}{imgs/scaling_recovery.pdf}
  \includegraphics[width=0.9\textwidth]{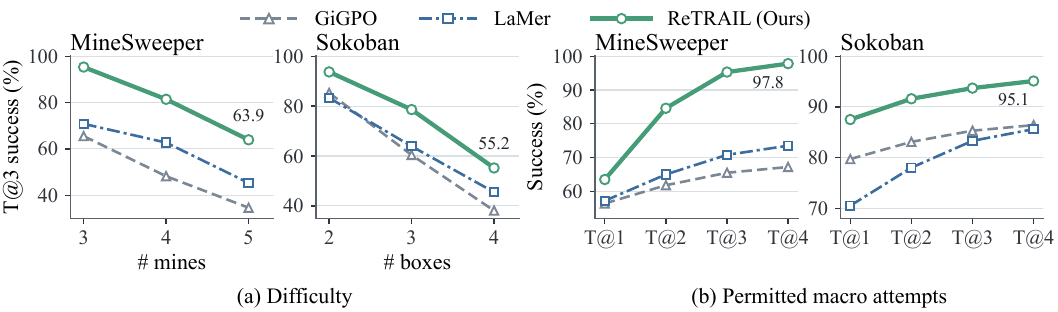}
  \caption{Scaling instance difficulty and recovery attempts.
  (a) T@3 success with additional mines or boxes; these settings vary instance difficulty.
  (b) Success with one to four permitted macro-attempts.
  ReTRAIL's endpoint value is annotated in each panel. Y-axis ranges differ by subplot and are
  explicitly labeled. Higher is better.}
  \label{fig:scaling_recovery}
  \vspace{-10pt}
\end{figure*}

\paragraph{Scaling to harder instances.}
We increase the number of mines in MineSweeper and boxes in Sokoban to construct harder variants. Figure~\ref{fig:scaling_recovery}(a) shows that success decreases for all plotted methods, while ReTRAIL retains an advantage at every displayed setting. At the hardest setting, ReTRAIL reaches \(63.9\) on MineSweeper, compared with \(45.3\) for the strongest baseline, and \(55.2\) on Sokoban, compared with \(45.5\), yielding margins of \(18.6\) and \(9.7\) percentage points. These results show that ReTRAIL maintains its advantage on harder instances. The analysis by shortest-solution length is reported separately in Figure~\ref{fig:stratified}.

\paragraph{Scaling the number of attempts.}
Allowing up to four macro-attempts provides additional opportunities for recovery with accumulated experience. ReTRAIL achieves the highest success rate throughout Figure~\ref{fig:scaling_recovery}(b). On MineSweeper, success increases from \(63.5\) at T@1 to \(97.8\) at T@4, a gain of \(34.3\) percentage points, compared with \(10.8\) for GiGPO and \(16.3\) for LaMer. On Sokoban, ReTRAIL reaches \(95.1\) at T@4 and remains highest at each attempt budget. Thus, the learned execution and reflection policy continues to benefit from additional recovery opportunities.

%% file: tables/main_results.tex
\begin{table*}[t]
  \centering
  \caption{Performance comparison on four long-horizon interactive benchmarks. \textbf{P@1}
  evaluates forward execution without rewind, \textbf{T@1} enables proactive rewind within one
  trial, and \textbf{T@3} permits up to three macro attempts with reflection-guided recovery.
  Every setting shares a 60-call budget per trial, including execution and reflection. Colored
  annotations show absolute changes relative to P@1; bold values indicate the best result in each
  column. All values are success rates (\%).}
  \label{tab:main_results}
  \vspace{0.25em}
  \begingroup
  \setlength{\tabcolsep}{2.7pt}
  \renewcommand{\arraystretch}{1.12}
  \small
  \resizebox{0.98\textwidth}{!}{%
  \begin{tabular}{l*{4}{ccc}}
    \toprule
    \multirow{2}{*}{\textbf{Method}}
      & \multicolumn{3}{c}{\textbf{MineSweeper}}
      & \multicolumn{3}{c}{\textbf{Sokoban}}
      & \multicolumn{3}{c}{\textbf{ALFWorld}}
      & \multicolumn{3}{c}{\textbf{ScienceWorld}} \\
    \cmidrule(lr){2-4} \cmidrule(lr){5-7} \cmidrule(lr){8-10} \cmidrule(lr){11-13}
      & \textbf{P@1} & \textbf{T@1} & \textbf{T@3}
      & \textbf{P@1} & \textbf{T@1} & \textbf{T@3}
      & \textbf{P@1} & \textbf{T@1} & \textbf{T@3}
      & \textbf{P@1} & \textbf{T@1} & \textbf{T@3} \\
    \midrule
    \multicolumn{13}{l}{\textit{Prompting Methods}} \vspace{0.35em} \\
    \rowcolor{black!3}
    Zero-shot
      & 3.5 & 3.1\gn{0.4} & 5.2\gp{1.7}
      & 6.0 & 5.5\gn{0.5} & 9.0\gp{3.0}
      & 35.0 & 34.2\gn{0.8} & 40.1\gp{5.1}
      & 8.0 & 7.2\gn{0.8} & 11.0\gp{3.0} \\
    ReAct
      & 6.0 & 6.5\gp{0.5} & 10.2\gp{4.2}
      & 7.0 & 6.8\gn{0.2} & 11.8\gp{4.8}
      & 40.0 & 41.0\gp{1.0} & 47.5\gp{7.5}
      & 10.2 & 9.4\gn{0.8} & 15.1\gp{4.9} \\
    \rowcolor{black!3}
    Reflexion
      & 5.5 & 7.0\gp{1.5} & 9.8\gp{4.3}
      & 6.5 & 6.2\gn{0.3} & 12.0\gp{5.5}
      & 38.0 & 42.1\gp{4.1} & 45.3\gp{7.3}
      & 9.0 & 13.0\gp{4.0} & 16.2\gp{7.2} \\
    Rollback
      & 6.7 & 6.5\gn{0.2} & 13.2\gp{6.5}
      & 9.7 & 10.3\gp{0.6} & 15.0\gp{5.3}
      & 35.0 & 34.2\gn{0.8} & 43.1\gp{8.1}
      & 13.2 & 12.3\gn{0.9} & 20.0\gp{6.8} \\
    \midrule
      \multicolumn{13}{l}{\textit{Fine-Tuning Methods}} \vspace{0.35em} \\
    \rowcolor{black!3}
    GRPO
      & 54.6 & 51.6\gn{3.0} & 60.8\gp{6.2}
      & 70.3 & 68.9\gn{1.4} & 82.8\gp{12.5}
      & 91.3 & 90.8\gn{0.5} & 93.5\gp{2.2}
      & 48.9 & 47.5\gn{1.4} & 52.1\gp{3.2} \\
    RLOO
      & 53.2 & 50.4\gn{2.8} & 59.1\gp{5.9}
      & 69.1 & 67.8\gn{1.3} & 81.0\gp{11.9}
      & 90.7 & 90.2\gn{0.5} & 92.8\gp{2.1}
      & 47.6 & 46.3\gn{1.3} & 50.8\gp{3.2} \\
    \rowcolor{black!3}
    GiGPO
      & 57.9 & 56.4\gn{1.5} & 65.5\gp{7.6}
      & \textbf{78.9} & 79.7\gp{0.8} & 85.3\gp{6.4}
      & 90.9 & 90.5\gn{0.4} & 94.3\gp{3.4}
      & 56.1 & 55.0\gn{1.1} & 58.3\gp{2.2} \\
    BEACON
      & 58.5 & 57.2\gn{1.3} & 68.0\gp{9.5}
      & 77.0 & 76.1\gn{0.9} & 84.0\gp{7.0}
      & 91.8 & 91.2\gn{0.6} & 94.5\gp{2.7}
      & 56.5 & 55.1\gn{1.4} & 61.0\gp{4.5} \\
    \rowcolor{black!3}
    LaMer
      & 55.4 & 57.2\gp{1.8} & 70.8\gp{15.4}
      & 72.1 & 70.5\gn{1.6} & 83.3\gp{11.2}
      & 89.7 & 90.1\gp{0.4} & 95.7\gp{6.0}
      & 53.5 & 54.2\gp{0.7} & 60.2\gp{6.7} \\
    \rowcolor{blue!8}
    \textbf{ReTRAIL}
      & \textbf{58.8} & \textbf{63.5}\gp{4.7} & \textbf{95.3}\gp{36.5}
      & 76.6 & \textbf{87.5}\gp{10.9} & \textbf{93.7}\gp{17.1}
      & \textbf{92.7} & \textbf{94.0}\gp{1.3} & \textbf{98.5}\gp{5.8}
      & \textbf{56.7} & \textbf{62.0}\gp{5.3} & \textbf{71.2}\gp{14.5} \\
    \bottomrule
  \end{tabular}%
  }
  \endgroup
  \vspace{-0.35em}
\end{table*}

%% file: sections/analysis.tex
\vspace{-1pt}

\section{Analysis}
\label{sec:analysis}


\input{tables/ablations_side_by_side}

\paragraph{Ablation.} Tables~\ref{tab:ablation_flow} and~\ref{tab:ablation_rl} evaluate the roles of retrospective exploration and its learning objectives.
Removing model-initiated rewind lowers T@1 from \(63.5\) to \(58.1\) on MineSweeper and from \(87.5\) to \(72.3\) on Sokoban, showing the benefit of proactive recovery.
Replacing distilled reflection with the raw abandoned suffix has a smaller effect on T@1, but lowers T@3 to \(69.7\) and \(87.2\), respectively.
This suggests that distilling branch experience provides more useful guidance for subsequent attempts than retaining raw history alone.

We next ablate the learning objective. 
Replacing the branch-advancement reward $r_m^{\mathrm{ref}} = v_m^+ - v_m^-$ with post-recovery progress $v_m^+$ alone lowers T@3 from \(95.3\) to \(80.2\) on MineSweeper and from \(93.7\) to \(84.9\) on Sokoban, alongside smaller declines in T@1.
These results support using subsequent branch progress to guide reflection.
We also replace State-Anchored GRPO with GiGPO within the same retrospective-exploration framework; this variant is distinct from the standalone GiGPO baseline in Table~\ref{tab:main_results}.
T@1/T@3 falls from \(63.5/95.3\) to \(57.6/90.1\) on MineSweeper and from \(87.5/93.7\) to \(84.2/90.3\) on Sokoban.
Together, these results support the proposed execution and reflection objectives, with the GiGPO comparison evaluating the credit-assignment designs as implemented.

\paragraph{Recovery Gains across Task Horizons.}
We stratify tasks by reference-solution length to examine how ReTRAIL's recovery gains vary with task horizon (Appendix~\ref{app:stratification}).
From short to long tasks, its reflection-guided retry gain increases from \(4.5\) to \(19.0\) percentage points, while its proactive gain rises from \(1.3\) to \(7.8\) points.
Although LaMer achieves a slightly larger retry gain on short tasks, ReTRAIL surpasses it by \(5.8\) points on long tasks.
These results suggest that retrospective exploration is particularly beneficial for long-horizon execution.

%% file: tables/ablations_side_by_side.tex
\begin{table}[t]
  \centering
  \resizebox{\textwidth}{!}{
  \def\PairedAblations{1}
  \newcommand{\AblationVariant}[3]{%
    \ifnum\RevisionMode=1\relax
      \shortstack[l]{\textcolor{RevisionOld}{#2}\\\textcolor{RevisionNew}{#3}}%
    \else\RevInline{#1}{#2}{#3}\fi}
  \ifnum\RevisionMode=1\relax
    \def\AblationCaptionHeight{3\baselineskip}
  \else
    \def\AblationCaptionHeight{1.5\baselineskip}
  \fi
  \begin{minipage}[t]{0.46\linewidth}
    \input{tables/ablation_flow}
  \end{minipage}\hspace{0.01\linewidth}
  \begin{minipage}[t]{0.50\linewidth}
    \input{tables/ablation_rl}
  \end{minipage}
  }
  \vspace{-1pt}
\end{table}

%% file: tables/ablation_flow.tex
\ifdefined\PairedAblations\else\ifnum\RevisionMode=2\relax\begin{wraptable}{r}{0.61\textwidth}\else\begin{table}[tbp]\fi\fi
  \centering
  \ifdefined\PairedAblations
    \begin{minipage}[t][\AblationCaptionHeight][t]{\linewidth}
  \caption{Ablation of retrospective exploration. Success rate (\%).}
  \label{tab:ablation_flow}
    \end{minipage}\par
  \else
  \caption{\RevInline{N14}{Ablation of the rollback-aware rollout.}{Ablation of retrospective exploration.} Success rate (\%); best per column in
  \textbf{bold}.}
  \label{tab:ablation_flow}
  \fi
  \vspace{1pt}
  \ifdefined\PairedAblations\setlength{\tabcolsep}{1.6pt}\else\setlength{\tabcolsep}{2.5pt}\fi
  \renewcommand{\arraystretch}{1.08}
\footnotesize
  \begin{tabularx}{\linewidth}{@{}lYYYY@{}}
    \toprule
    \multirow{2}{*}{\textbf{Variant}}
      & \multicolumn{2}{c}{\textbf{MineSweeper}}
      & \multicolumn{2}{c}{\textbf{Sokoban}} \\
    \cmidrule(lr){2-3}\cmidrule(lr){4-5}
      & T@1 & T@3 & T@1 & T@3 \\
    \midrule
    \rowcolor{black!8}
    ReTRAIL (full)
      & \textbf{63.5} & \textbf{95.3} & \textbf{87.5} & \textbf{93.7} \\
    \ifdefined\PairedAblations\hspace{0.35em}\else\quad\fi w/o model rewind
      & 58.1 & 78.4 & 72.3 & 89.1 \\
    \ifdefined\PairedAblations\hspace{0.35em}\else\quad\fi \shortstack[l]{w/o reflection}
      & 61.8 & 69.7 & 86.0 & 87.2 \\
 
    \bottomrule
  \end{tabularx}
\ifdefined\PairedAblations\else\ifnum\RevisionMode=2\relax\end{wraptable}\else\end{table}\fi\fi

%% file: tables/ablation_rl.tex
\ifdefined\PairedAblations\else\ifnum\RevisionMode=2\relax\begin{wraptable}{r}{0.61\textwidth}\else\begin{table}[tbp]\fi\fi
  \centering
  \ifdefined\PairedAblations
    \begin{minipage}[t][\AblationCaptionHeight][t]{\linewidth}
  \caption{Ablation of the learning objective. Success rate (\%).}
  \label{tab:ablation_rl}
    \end{minipage}\par
  \else
  \caption{Ablation of the learning objective. Success rate (\%).}
  \label{tab:ablation_rl}
  \fi
  \vspace{1pt}
  \ifdefined\PairedAblations\setlength{\tabcolsep}{1.6pt}\else\setlength{\tabcolsep}{2.5pt}\fi
  \renewcommand{\arraystretch}{1.08}
  \footnotesize
  \begin{tabularx}{\linewidth}{@{}lYYYY@{}}
    \toprule
    \multirow{2}{*}{\textbf{Variant}}
      & \multicolumn{2}{c}{\textbf{MineSweeper}}
      & \multicolumn{2}{c}{\textbf{Sokoban}} \\
    \cmidrule(lr){2-3}\cmidrule(lr){4-5}
      & T@1 & T@3 & T@1 & T@3 \\
    \midrule
    \rowcolor{black!8}
    ReTRAIL (full)
      & \textbf{63.5} & \textbf{95.3} & \textbf{87.5} & \textbf{93.7} \\
    \ifdefined\PairedAblations\hspace{0.35em}\else\quad\fi w/o progress-diff reward
      & 60.9 & 80.2 & 81.7 & 84.9 \\
    \ifdefined\PairedAblations\hspace{0.35em}\else\quad\fi GiGPO objective
      & 57.6 & 90.1 & 84.2 & 90.3 \\
 
    \bottomrule
  \end{tabularx}
\ifdefined\PairedAblations\else\ifnum\RevisionMode=2\relax\end{wraptable}\else\end{table}\fi\fi

%% file: sections/conclusion.tex
\vspace{-2pt}

\section{Conclusion}
\label{sec:conclusion}

Learning to recover from unproductive exploration is crucial for autonomous agents to make sustained progress on long-horizon tasks. We introduced \textbf{ReTRAIL}, a framework that combines execution recovery with experiential learning through retrospective exploration. With an environment-level \textit{rewind} interface, agents can revisit earlier states and explore alternative continuations while retaining experience from abandoned branches. ReTRAIL trains a shared execution and reflection policy under this protocol, using state-anchored comparisons to improve execution and branch-advancement feedback to guide reflection. Across four interactive benchmarks, ReTRAIL consistently outperforms both methods without rewind and baselines equipped with the same interface, with ablations supporting the complementary roles of recovery, retained experience, and progress-based feedback. More broadly, these findings highlight retrospective exploration as a promising foundation for autonomous agents that can recover from setbacks, adapt through accumulated experience, and sustain progress toward long-horizon goals.

%% file: sections/ai_statement.tex
\section*{AI Use Statement}
In preparing this manuscript, we used generative AI tools for language polishing and literature search and survey. Language assistance focused on improving grammar, sentence flow, clarity, and readability while preserving the intended technical meaning. Literature assistance supported the identification of relevant prior work and the organization and summarization of existing research. The authors remain responsible for reviewing all AI-assisted material, checking literature summaries and citations against the original sources, and ensuring that the manuscript accurately represents the research. We take full responsibility for the final content of this work, including all text, claims, and artifacts produced with the assistance of generative AI.

%% file: sections/ethics_statement.tex
\section*{Ethics Statement}
This work studies recovery and learning in autonomous agents through experiments on four controlled benchmark environments: MineSweeper, Sokoban, ALFWorld, and ScienceWorld. The framework aims to improve agents' ability to recover from unproductive exploration, but state restoration in these environments should not be interpreted as reversing the consequences of actions in real-world systems. Deployment beyond these benchmarks should therefore include appropriate authorization, safeguards for external actions, and human oversight where actions may have consequential or irreversible effects. Retained interaction histories and reflection memories should be handled with appropriate privacy and access controls, and any reuse or release of data, models, and software should respect their applicable licenses and terms of use.

%% file: sections/reproducibility_statement.tex
\section*{Reproducibility Statement}
To support reproducibility, we specify the retrospective-exploration framework and learning objectives in Sections~\ref{sec:autonomous-execution} and~\ref{sec:policy_optimization}. Appendix~\ref{app:method_details} details environment restoration, recovery and context management, and provides pseudocode for rollout collection and policy updates. Section~\ref{sec:exp-setup} and Appendix~\ref{app:experimental_setup} document the benchmarks and data partitions, baseline configurations, training hyperparameters, reward and progress signals, evaluation protocols, model-call accounting, and aggregation across benchmarks and seeds. Appendix~\ref{app:prompts} provides execution and reflection prompt templates and examples of observation formatting for the four benchmark environments.

%% file: sections/appendix.tex
\clearpage
\input{sections/appendix/discussion}
\input{sections/appendix/additional_results}
\input{sections/appendix/experimental_setup}

\input{sections/appendix/method_details}

\input{sections/prompts}
\clearpage
\input{sections/case_study}

%% file: sections/appendix/discussion.tex
\section{Limitations and Future Work}
\label{app:discussion}
\label{app:scope}

Our results motivate two directions for extending retrospective exploration.

\paragraph{(i) Extending recovery to broader environments.}
The current framework assumes access to state restoration through snapshots or
reset-and-replay; the latter requires deterministic transitions and resettable
external state (Appendix~\ref{app:restoration}). Extending recovery to environments
with approximate restoration or irreversible side effects is an important next
step. Future work could explore restoration controllers and compensating actions,
together with checks on the recovered state, to determine when earlier progress
can be safely reused.

\paragraph{(ii) Improving recovery credit and efficiency.}
ReTRAIL combines task-wide execution feedback with branch advancement, an
environment-progress proxy for the utility of retained experience. The current
execution return ends at each rewind boundary, while subsequent advancement
supervises reflection (Appendix~\ref{app:credit_example}). This design motivates
studying credit assignment~\citep{wang2026agentopsdrecursiveselfdistillationagentic,wang2026castgamesolversturnlevel} across recovery boundaries, so that rewind timing and
target selection can receive direct feedback from the resumed branch. Accounting
for restoration and reflection costs, and developing progress estimators for
settings with sparse or noisy feedback, could further broaden the framework's
practical use.

%% file: sections/appendix/additional_results.tex
\section{Additional Results and Analysis}
\label{app:additional_results}

\subsection{Recovery Gains by Reference-Solution Length}
\label{app:stratification}
\label{sec:stratified}
\label{sec:behavior}

We group held-out instances by reference-solution length from the initial state.
All methods use the same groups, independent of their observed rollout lengths.

\paragraph{Reference-solution length and grouping.}
For Sokoban, the reference is the breadth-first-search solution length \(d_0\)
(Appendix~\ref{app:progress_signals}); for MineSweeper, it is the minimum reveal count under
correct deduction. For ALFWorld and ScienceWorld, it is the length of the provided expert or
gold action sequence. We use the term \emph{reference-solution length} because expert
sequences need not be shortest solutions. On ALFWorld, we follow the length boundaries
of~\citet{wang2026beacon}: Short (at most 4 actions), Medium (5--7), and Long (more than 7).
For each remaining environment, we use the terciles of its held-out reference lengths.
The bins therefore describe length within each environment, rather than a common absolute
action count across environments.

\paragraph{Metrics and aggregation.}
Within each length bin, we compute the proactive gain
\(\Delta_{\mathrm{pro}}=\mathrm{T@1}-\mathrm{P@1}\) and the reflection-guided retry gain
\(\Delta_{\mathrm{ref}}=\mathrm{T@3}-\mathrm{T@1}\) on the same instances for every method.
We then average each gain across the four environments with equal weight; values are
success-rate differences in percentage points. 

\begin{figure}[t]
  \centering
  \includegraphics[width=0.80\textwidth]{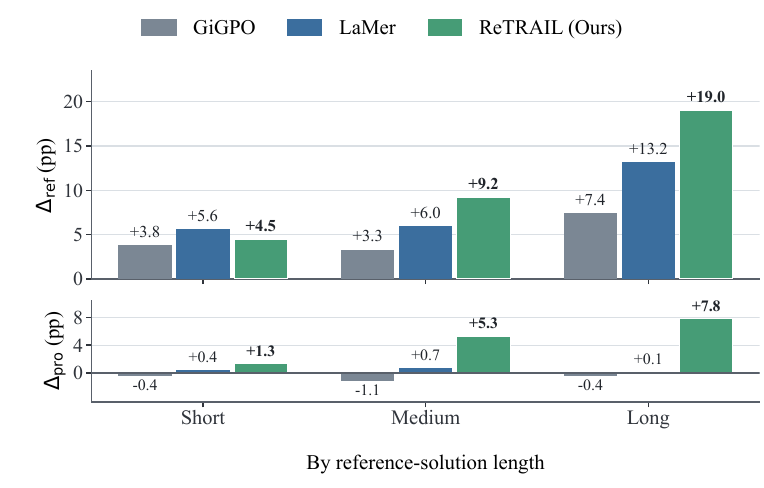}
  \caption{Recovery gains by reference-solution length. The upper row shows
  reflection-guided retry gain \(\Delta_{\mathrm{ref}}=\mathrm{T@3}-\mathrm{T@1}\);
  the lower row shows proactive gain \(\Delta_{\mathrm{pro}}=\mathrm{T@1}-\mathrm{P@1}\).
  Values are equal-weight averages across the four environments, in percentage points.
  Short, Medium, and Long use the environment-specific grouping rules above.}
  \label{fig:stratified}
\end{figure}

\paragraph{Observed pattern.}
ReTRAIL's retry gain increases from \(+4.5\) on short tasks to \(+9.2\) on medium and
\(+19.0\) on long tasks; LaMer's corresponding gains are \(+5.6\), \(+6.0\), and \(+13.2\).
Thus, ReTRAIL's retry-gain advantage over LaMer is \(-1.1\), \(+3.2\), and \(+5.8\) points
in the three groups. Its proactive gain also increases from \(+1.3\) to \(+5.3\) and \(+7.8\).
These descriptive results show larger recovery gains for ReTRAIL on longer-reference tasks,
while LaMer has the larger retry gain in the short group. They do not isolate length from
other task properties or establish a causal effect of reflection alone.

\subsection{Average Performance Across Benchmarks}

Table~\ref{tab:benchmark_aggregate} summarizes Table~\ref{tab:main_results} using
the equal-benchmark rule in Appendix~\ref{app:seed_aggregation}. It is computed
from the reported, rounded benchmark means, so minor rounding differences are
possible. The reported spread is across evaluation seeds, not additional runs.
The forward-only results support competitive performance, without
establishing statistically significant superiority on each benchmark.

\input{tables/seed_aggregation}

%% file: tables/seed_aggregation.tex
\providecommand{\SeedSDPlaceholder}{0}
\providecommand{\sd}[1]{${\pm}#1$\ifnum\SeedSDPlaceholder=1\relax\textsuperscript{\textcolor{red}{$\dagger$}}\fi}

\begin{table}[htbp]
\centering
\caption{Equal-weight averages across the four benchmarks, computed from the
reported benchmark means in Table~\ref{tab:main_results} and rounded to one
decimal place. \(\pm\) is the standard deviation across the three evaluation
seeds. These are aggregate point estimates, not additional runs.}
\label{tab:benchmark_aggregate}
\small
\setlength{\tabcolsep}{4.5pt}
\begin{tabular}{lccc}
\toprule
Method & P@1 & T@1 & T@3 \\
\midrule
Zero-shot & 13.1\,\sd{1.6} & 12.5\,\sd{1.7} & 16.3\,\sd{2.7} \\
ReAct & 15.8\,\sd{2.9} & 15.9\,\sd{1.2} & 21.2\,\sd{3.2} \\
Reflexion & 14.8\,\sd{2.5} & 17.1\,\sd{2.6} & 20.8\,\sd{1.4} \\
Rollback & 16.2\,\sd{2.7} & 15.8\,\sd{1.7} & 22.8\,\sd{3.1} \\
GRPO & 66.3\,\sd{1.2} & 64.7\,\sd{1.4} & 72.3\,\sd{2.0} \\
RLOO & 65.2\,\sd{2.3} & 63.7\,\sd{2.3} & 70.9\,\sd{2.1} \\
GiGPO & 71.0\,\sd{1.7} & 70.4\,\sd{2.0} & 75.9\,\sd{1.9} \\
BEACON & 71.0\,\sd{2.5} & 69.9\,\sd{1.1} & 76.9\,\sd{2.0} \\
LaMer & 67.7\,\sd{0.9} & 68.0\,\sd{2.2} & 77.5\,\sd{2.4} \\
\textbf{ReTRAIL} & \textbf{71.2}\,\sd{1.5} & \textbf{76.8}\,\sd{1.9} & \textbf{89.7}\,\sd{2.2} \\
\bottomrule
\end{tabular}

\end{table}

%% file: sections/appendix/experimental_setup.tex
\section{Experimental Setup}
\label{app:experimental_setup}

\subsection{Benchmarks and Data Partitions}
\label{app:env_details}

All four benchmarks use text interaction and an environment-restoration interface.
Table~\ref{tab:dataset_partitions} lists the registered task instances loaded by
the training entrypoints. The registered Sokoban variants use $6\times6$ or
$7\times7$ rooms with two or three boxes; each has 256 training and 128 test
entries. The difficulty experiment varies the box count separately.
Dataset counts refer to task entries, not repeated rollouts or evaluation seeds.

\input{tables/dataset_partitions}

\subsection{Rewards and Progress Signals}
\label{app:progress_signals}

The optimizer uses three quantities: the per-decision reward \(\delta_t\),
the segment-endpoint score \(S\), and the branch score \(V(\xi\mid s)\).
Table~\ref{tab:progress_signals} gives their definitions. If one decision executes
several primitive actions, \(\delta_t\) sums their environment rewards. A decision
with no environment transition, including a rewind request, has \(\delta_t=0\).

\begin{table}[t]
  \centering
  \caption{Training signals by environment. \(\delta_t\) supplies the execution
  return, \(S\) the task-wide endpoint score, and \(V\) the branch score for reflection.
  Derived advantages and branch-score differences are not policy inputs.}
  \label{tab:progress_signals}
  \small
  \setlength{\tabcolsep}{4pt}
  \begin{tabularx}{\textwidth}{@{}l X X X@{}}
    \toprule
     & \(\delta_t\) (per step) & \(S\) (segment endpoint) & \(V(\xi\mid s)\) (branch) \\
    \midrule
    MineSweeper & native board reward: \(+0.5\) safe reveal, \(+2.0\) blank-region expansion,
    \(+10.0\) board cleared, \(-0.1\) mine revealed, \(-1.0\) invalid or no-change action
     & cumulative \(\delta\) on the retained path from \(s_0\) & cumulative \(\delta\) along the branch \\
    \addlinespace[2pt]
    Sokoban & environment-returned transition reward & \(d_0-d(\text{endpoint})\) by breadth-first
    search & \(d_0-d_{\mathrm{end}}\), with \(d_{\mathrm{end}}=0\) if solved and
    \(d_{\mathrm{end}}=d_0\) if deadlocked \\
    \addlinespace[2pt]
    ALFWorld & environment-returned transition reward & fraction of satisfied goal conditions at the
    endpoint & maximum fraction of satisfied goal conditions along the branch \\
    \addlinespace[2pt]
    ScienceWorld & environment-returned transition reward & native completion score at the endpoint &
    maximum native completion score along the branch \\
    \midrule
    \emph{Policy visibility} & environment-dependent native feedback & ScienceWorld score is visible & the branch comparison is not a prompt input \\
    \bottomrule
  \end{tabularx}
\end{table}

\paragraph{MineSweeper.}
Completion and mine revelation end the attempt, with the corresponding
terminal reward in Table~\ref{tab:progress_signals} applied to that action.
\(S\) accumulates rewards along the retained path from \(s_0\); restoring a
checkpoint also restores this prefix score. \(V\) sums rewards only after the
selected anchor, so the common prefix cancels in the branch comparison.

\paragraph{Sokoban.}
Let \(d_0\) be the shortest-solution distance from the initial state and
\(d_{\mathrm{end}}\) the remaining distance at a branch endpoint, both computed
by breadth-first search. We use \(V=d_0-d_{\mathrm{end}}\), taking
\(d_{\mathrm{end}}=0\) for a solved state and \(d_{\mathrm{end}}=d_0\) for a
deadlocked or unsolvable state. The new-minus-old branch score is therefore
\(d_{\mathrm{end}}^{-}-d_{\mathrm{end}}^{+}\). These search distances are
evaluator-only and are not shown to the policy.

\paragraph{ALFWorld and ScienceWorld.}
ALFWorld measures the fraction of satisfied goal conditions; ScienceWorld uses
the native task-completion score. In each environment, \(S\) is the endpoint
score and \(V\) is the maximum score along the branch, including its starting
state. These differ when progress can be undone. ScienceWorld's native score
is visible in its observations; the derived branch-advancement reward is not.

\subsection{Training Configuration and Sampling Budget}
\label{app:training_details}
\label{app:hyperparameters}

All trained methods start from Qwen3-4B~\citep{yang2025qwen3}. They share the
optimizer, learning-rate schedule, task splits, and decoding settings.
Table~\ref{tab:hyperparameters} lists the configuration. Prompt serialization
follows each rollout interface (Appendix~\ref{app:prompts}); shared task
instructions do not imply identical execution and reflection conversations.

Per task and update, baselines sample 24 trajectories, each allowing 60 model
calls. ReTRAIL samples eight trials, each allowing 60 calls and at most three
macro-attempts. This aligns nominal attempt slots, with call ceilings of
1,440 for the baseline and 480 for ReTRAIL. These ceilings do not measure
realized calls, tokens, or compute. Experience and advantage-normalization
groups remain within each task.

\input{tables/hyperparameters}

\subsection{Evaluation Protocol and Call Accounting}
\label{app:control_flow}

Table~\ref{tab:appendix_units} distinguishes the units used throughout the paper.
Rewinding restores an active path but never refunds calls already made.

\begin{table}[t]
\centering
\caption{Units used for rollout and budget accounting.}
\label{tab:appendix_units}
\small
\begin{tabularx}{\textwidth}{@{}l X@{}}
\toprule
Unit & Meaning \\
\midrule
Trial & The full run under one global model-call budget. \\
Macro-attempt & One attempt until success, irreversible failure, or budget exhaustion; proactive rewinds stay within it. \\
Forward segment & Execution between recovery boundaries, or until success, failure, or budget exhaustion. \\
Model call & One execution or reflection generation; a rewind request is an execution call. \\
Primitive action & One environment action; a Sokoban execution call can contain several moves. \\
\bottomrule
\end{tabularx}
\end{table}

P@1 disables recovery. T@1 allows proactive recovery within one macro-attempt
and stops at the first irreversible failure. T@3 also permits failure-triggered
recovery, up to three macro-attempts. Training uses the same attempt cap.
Model-initiated rewinds are limited to three per trial. The attempt-scaling
experiment varies the macro-attempt cap from one to four.

Training and evaluation impose a global ceiling of 60 model calls per trial:
\(p+f\le60\), where \(p\) counts execution and \(f\) reflection.
A proactive recovery uses an execution call to request rewind and a separate
reflection call. Forced recovery uses a reflection call after the preceding
execution. Recovery starts only when its attempt/rewind allowance permits it
and at least two calls remain for reflection and continuation.
Reset-and-replay without model generation adds environment actions but no model calls.

The prompt backends also track a branch allowance. MineSweeper and Sokoban
restore its saved consumption at the checkpoint; the supplied ALFWorld flow
resets its segment-turn counter. These backend rules are distinct from the
global call ceiling. A forward-segment cap counts both recovery types,
whereas the macro-attempt count increases only after failure-triggered recovery.
Algorithm~\ref{alg:rollout} takes these rules as configuration inputs.

\subsection{Aggregation Across Benchmarks and Seeds}
\label{app:seed_aggregation}

Let \(p_{s,e}^{(k)}\) be success on benchmark \(e\), under evaluation seed \(s\)
and protocol \(k\in\{\mathrm{P@1},\mathrm{T@1},\mathrm{T@3}\}\). We average
equally across the four benchmarks and three seeds:
\begin{equation}
 M_s^{(k)}=\frac14\sum_{e=1}^{4}p_{s,e}^{(k)},\qquad
 \overline M^{(k)}=\frac13\sum_{s=1}^{3}M_s^{(k)}
 =\frac14\sum_{e=1}^{4}\overline p_e^{(k)}.
 \label{eq:seed_benchmark_aggregation}
\end{equation}
Each benchmark has equal weight regardless of its instance count. This differs
from pooling all test instances. Table~\ref{tab:benchmark_aggregate} reports the
resulting means together with the standard deviation across the three evaluation
seeds. Individual seed values cannot be recovered from these means.

%% file: tables/dataset_partitions.tex
\begin{table}[htbp]
\centering
\caption{Registered task-instance partitions loaded by the training entrypoints. Counts describe distinct dataset entries, not generated continuations or repeated evaluations. ALFWorld's selected test partition is \texttt{valid\_seen}; its separately registered \texttt{valid\_unseen} partition contains 134 entries. ScienceWorld uses the L0 partition.}
\label{tab:dataset_partitions}
\small
\begin{tabular}{lrrl}
\toprule
Environment & Train & Test & Partition information \\
\midrule
MineSweeper & 256 & 128 & $6\times6$ boards, 3 mines \\
Sokoban & 256 & 128 & registered room variants share these sizes \\
ALFWorld & 3{,}553 & 140 & \texttt{train} / \texttt{valid\_seen} \\
ScienceWorld & 3{,}322 & 1{,}661 & L0 task/variation pairs \\
\bottomrule
\end{tabular}
\end{table}

%% file: tables/hyperparameters.tex
\begin{table}[t]
  \centering
  \caption{Training and evaluation configuration of ReTRAIL. Optimizer settings are shared with the trained baselines. The retrospective-exploration
  and objective blocks describe ReTRAIL; a trial can contain several forward segments
  and should not be equated with one baseline trajectory.}
  \label{tab:hyperparameters}
  \vspace{1pt}
  \setlength{\tabcolsep}{4pt}
  \renewcommand{\arraystretch}{1.06}
  \footnotesize
  \begin{tabularx}{\linewidth}{@{}>{\raggedright\arraybackslash}Xl@{}}
    \toprule
    \textbf{Setting} & \textbf{Value} \\
    \midrule
    \multicolumn{2}{@{}l}{\textit{Policy and optimization}} \\
    Learning rate & \(1\times10^{-6}\) \\
    LR Scheduler & Constant \\
    Optimizer updates & 300 \\
    Batch size & 16 \\
    Advantage-normalization \(\epsilon\) & \(1\times10^{-6}\) \\
    Discount \(\gamma\) & 1.0 \\
    \midrule
    \multicolumn{2}{@{}l}{\textit{Retrospective exploration rollout}} \\
    Baseline trajectories per task/update & 24 \\
    ReTRAIL trials per task/update \(B\) & 8 \\
    Macro-attempts per trial (training) & at most 3 \\
    Model calls per ReTRAIL trial, execution + reflection & 60 \\
    Model calls per baseline trajectory & 60 \\
    Maximum calls per task/update, baseline / ReTRAIL & 1440 / 480 \\
    Model-initiated rewinds per trial & at most 3 \\
    Context length & 16{,}384 \\
    \midrule
    \multicolumn{2}{@{}l}{\textit{Objective weights}} \\
    State-anchored weight \(\lambda_{\mathrm{state}}\) & 0.5 \\
    Reflection weight \(\lambda_{\mathrm{ref}}\) & 0.5 \\
    \midrule
    \multicolumn{2}{@{}l}{\textit{Evaluation}} \\
    Evaluation seeds & 3, reported as the mean \\
    \bottomrule
  \end{tabularx}
\end{table}

%% file: sections/appendix/method_details.tex
\section{Method Details}
\label{app:method_details}

This section follows the data flow: restore an environment state, resume execution
with branch memory, record the rollout, and construct the training objective.

\subsection{Environment Restoration and Checkpoints}
\label{app:restoration}

The environment harness implements rewind. MineSweeper and Sokoban store and
restore serializable state snapshots. ALFWorld and ScienceWorld reset the task
and replay the recorded actions up to the selected point, since their engines
do not expose a general snapshot interface. Replay requires deterministic
transitions and resettable external state.

The active path stores ordered checkpoints \(\mathcal K=(C_0,\ldots,C_k)\),
each containing its state, path history, and restoration information. A target
\(j<k\) identifies a path position: identical observations can occur at different
positions. Restoring \(C_j\) truncates the active path but preserves the chronological
training archive \(\Gamma\). This archive stores all outputs with their original
input contexts, segment boundaries, feedback, and abandoned/resumed branch pairs.
Thus an abandoned branch remains available for training after leaving the live prompt.

\subsection{Recovery Triggers and Reflection}
\label{app:recovery_mechanism}

A model-initiated recovery begins with an execution output requesting rewind.
A forced recovery begins when the controller detects a configured failure.
Both generate reflection before restoration, while the abandoned branch is
still available. The reflection returns branch memory in \texttt{<remark>} tags.

For MineSweeper and Sokoban, a valid model-selected target uses a known-target
reflection prompt; forced recovery asks the model to select a checkpoint as
well. ALFWorld and ScienceWorld also request a checkpoint in their reflection
responses. Thus \emph{forced} identifies who triggers recovery, not a fixed
destination immediately before failure. Table~\ref{tab:recovery_paths} summarizes
the two paths; Appendix~\ref{app:control_flow} specifies their budget limits.

\begin{table}[htbp]
  \centering
  \caption{\textbf{Recovery triggers and model calls.} Restoration follows reflection, so the
  abandoned branch remains available when its memory is generated. Checkpoint selection and
  fallback parsing follow the environment-specific prompt implementation.}
  \label{tab:recovery_paths}
  \small
  \setlength{\tabcolsep}{4pt}
  \begin{tabularx}{\textwidth}{@{}l X X@{}}
    \toprule
     & Model-initiated recovery & Forced recovery \\
    \midrule
    Trigger & execution output requests an earlier checkpoint
     & controller observes a configured failure condition \\
    \addlinespace[2pt]
    Target & requested by the policy; validated before restoration
     & selected through the checkpoint-choice reflection interface \\
    \addlinespace[2pt]
    Reflection & a separate generation returning \texttt{<remark>}
     & a separate generation returning \texttt{<remark>} and a target \\
    \addlinespace[2pt]
    Call accounting & one execution call for the request, plus one reflection call
     & one reflection call in addition to preceding execution calls \\
    \addlinespace[2pt]
    Main-text metric & available under T@1 and T@3
     & failure-triggered retries available under T@3 \\
    \bottomrule
  \end{tabularx}
\end{table}

\subsection{Context After Recovery}
\label{app:prompt_example}

The logical history \(h\) is the retained execution path. The actual input
\(c\) is the conversation produced by the environment's message builder.
The distinction matters because the backends serialize history differently.

MineSweeper and Sokoban save a conversation at each checkpoint. Recovery
restores this prefix, removes the abandoned continuation, and appends a user
message with the new branch memory. A previous trailing reflection at the
same checkpoint is first made available to the reflection call, then replaced.
The next observation is appended as another user message.

ScienceWorld also restores a message prefix, but includes branch memories in
its subsequent observation. ALFWorld rebuilds the resumed conversation from
the system message, restored observation, admissible commands, and branch
memories; it does not copy the earlier conversation prefix.
Appendix~\ref{app:prompts} gives the literal templates and formatting examples.
Training always uses the recorded input \(c\), without adding later reflections
to earlier samples.

\subsection{Rollout Algorithm}

Algorithm~\ref{alg:rollout} records one trial under a frozen behavior policy.
Its four blocks follow Section~\ref{sec:autonomous-execution}.
The active history \(h\), checkpoints \(\mathcal K\), and memory \(\mathcal E\)
guide execution; the archive \(\Gamma\) retains every generation with its
original context and feedback. A segment \(\sigma\) groups execution decisions
between recovery boundaries, and \(n=p+f\) counts all model calls.

Invalid outputs include unparseable actions and disabled or out-of-range rewind
requests; they consume a call and receive feedback without changing the environment.
Checkpoints include states reached on failure.
Recovery requires allowance for reflection and continuation
(Appendix~\ref{app:control_flow}). Target parsing follows
Appendix~\ref{app:recovery_mechanism}; memory and context updates follow
Appendix~\ref{app:prompt_example}. Restoring a checkpoint never truncates
\(\Gamma\) or refunds calls.

\begin{algorithm}[htbp]
  \caption{Retrospective exploration for one task}
  \label{alg:rollout}
  \small
  \begin{algorithmic}[1]
    \Require task \(u\), frozen policy \(\pi_{\mathrm{old}}\), protocol \(\mathcal B\), call limit \(C_{\max}=60\)
    \State reset task; initialize \(h,\mathcal K=(C_0),\mathcal E=\varnothing\) and protocol counters
    \State \(\Gamma\leftarrow\varnothing\); \(\sigma\leftarrow\varnothing\); \(n\leftarrow0\)
    \While{\(n<C_{\max}\) and \(\mathcal B\) permits execution}
      \Statex \hspace{\algorithmicindent}\textbf{1. Execute} \hfill Equation~\ref{eq:experience_conditioned_execution}
      \State \(c\leftarrow\textsc{BuildContext}(u,h,\mathcal E,\mathcal K)\);
        \(y\sim\pi_{\mathrm{old}}(\cdot\mid c)\); \(n\leftarrow n+1\)
      \State record \((c,y)\) in \(\Gamma\); append the execution decision to \(\sigma\)
      \State \textbf{if} \(y\) is invalid \textbf{then} record feedback; \textbf{continue}

      \Statex \hspace{\algorithmicindent}\textbf{2. Detect recovery} \hfill Equation~\ref{eq:rewind_partition}
      \If{\(y\) requests rewind}
        \State retain the requested target \Comment{model-initiated recovery}
      \Else
        \State execute the parsed action; update \(h,\mathcal K\); record rewards and observations in \(\Gamma\)
        \State \textbf{if} task solved \textbf{then break}
        \State \textbf{if} no failure trigger \textbf{then continue} \Comment{otherwise, forced recovery}
      \EndIf
      \State close \(\sigma\) in \(\Gamma\)
      \State \textbf{if} \(n+2>C_{\max}\) or \(\mathcal B\) disallows recovery/continuation \textbf{then break}

      \Statex \hspace{\algorithmicindent}\textbf{3. Reflect} \hfill Equation~\ref{eq:branch_experience}
      \State build \(c_{\mathrm{ref}}\) from the active branch, memory, and any requested target
      \State \(z\sim\pi_{\mathrm{old}}(\cdot\mid c_{\mathrm{ref}})\); \(n\leftarrow n+1\); record \((c_{\mathrm{ref}},z)\) in \(\Gamma\)
      \State resolve \((j,e)\) using any requested target, \(z\), and the configured fallback
      \State \textbf{if} no valid target \(0\le j<|\mathcal K|-1\) remains \textbf{then break}

      \Statex \hspace{\algorithmicindent}\textbf{4. Restore and resume} \hfill Equation~\ref{eq:rewind_resume}
      \State record abandoned suffix \(\beta=\tau_{j:k}\) and recovery link in \(\Gamma\), where \(k=|\mathcal K|-1\)
      \State restore \(C_j\); truncate \(h,\mathcal K\) to the retained prefix
      \State update \(\mathcal E\) and resumed context with \(e\); update branch/protocol counters
      \State start a new empty \(\sigma\) linked to this recovery
    \EndWhile
    \State close any nonempty open segment in \(\Gamma\); \Return \(\Gamma\)
  \end{algorithmic}
\end{algorithm}

\subsection{State-Anchored GRPO}
\label{app:state_anchored_grpo}

Each execution decision is one training sample. Its remaining segment supplies
the return used to score that decision, while the loss applies only to its own
generated output.

\subsubsection{Decision Samples and Returns}

For task \(u\), collect \(B\) trials \(\{\Gamma_i\}_{i=1}^{B}\) and their
forward segments \(\mathcal S(u)\). Index decision \(t\) in segment
\(\sigma_{i,m}\) by \(q=(i,m,t)\), and let \(T_{i,m}\) be the segment length:
\begin{equation}
    \mathcal I(u)
    =\{(i,m,t):\sigma_{i,m}\in\mathcal S(u),\;0\leq t<T_{i,m}\},
    \qquad
    \xi_q=\sigma_{i,m}[t:T_{i,m}].
    \label{eq:anchored_continuations}
\end{equation}
The continuation \(\xi_q\) ends with its original segment, at success, failure,
rewind, or budget exhaustion. A proactive rewind is the segment's final execution
decision. Record its starting state \(s_q\), logical retained history \(h_q\),
and actual input context. Let \(\mathcal Y_q\) contain only the generated
reasoning and action or rewind request for decision \(q\).
The indexed collection \(\mathcal C(u)=(\xi_q)_{q\in\mathcal I(u)}\) contains
return records, not extra sequences to train again.

For example, an old segment \(s_0\to s_1\to s_2\to s_3\) that ends in a rewind
to \(s_1\) supplies a continuation from \(s_1\), even though the segment began
at \(s_0\). This can be compared with the new continuation from \(s_1\).
The two decisions share a state and logical prefix, but retain their original
generation contexts and potentially different branch memories.

\subsubsection{Execution Advantages}

The return from decision \(q\) and its comparison group are
\begin{equation}
    G_q=\sum_{\ell=t}^{T_{i,m}-1}\gamma^{\ell-t}\delta_{i,m,\ell},
    \qquad
    \mathcal H_u(s,h)=\{q\in\mathcal I(u):s_q=s,\;h_q=h\},
    \label{eq:state_anchor_group}
\end{equation}
Here \(\delta_{i,m,\ell}\) sums environment rewards produced by execution
decision \(\ell\). A rewind request has zero transition reward; reflection
calls do not enter this return. Group members share a task, environment state,
and retained history. Their accumulated experience and remaining global budget
may differ, so the comparison does not hold the full policy input fixed.

The macro signal uses endpoint progress \(S_{i,m}\), with each segment
contributing once to normalization regardless of its length. For any decision
in that segment, the main text's \(S_{\xi_q}\) equals \(S_{i,m}\). The combined
advantage is
\begin{equation}
\begin{aligned}
    A_q^{\mathrm{state}}
    &=\operatorname{Norm}_{\mathcal H_u(s_q,h_q)}(G_q),\\
    A_q^{\mathrm{cont}}
    &=\underbrace{\operatorname{Norm}_{\mathcal S(u)}(S_{i,m})}_{A_{i,m}^{\mathrm{macro}}}
      +\lambda_{\mathrm{state}}A_q^{\mathrm{state}}.
\end{aligned}
\label{eq:state_anchored_advantage_details}
\end{equation}
\(\operatorname{Norm}\) subtracts the group mean and divides by its standard
deviation plus \(\epsilon\). Singleton and constant-score groups have zero
advantage; empty groups contribute no samples. These rules apply separately
to macro, state, and reflection groups within each task. When a decision has
no local comparison, its macro term remains available.

All decisions in a segment share its macro advantage but have their own local
advantages. Each advantage applies only to that decision's output. A reward
can contribute to several earlier returns without duplicating output tokens
in the loss. Repeated visits are separate generation events.

\subsubsection{Reflection Rewards and Eligibility}

At a recovery point, reflection is scored by the improvement of the resumed
branch over the abandoned suffix:
\begin{equation}
    r_m^{\mathrm{ref}} = V(\sigma_{m+1}\mid s_{j_m}) - V(\beta_m\mid s_{j_m}).
    \label{eq:appendix_reflection_reward_discussion}
\end{equation}
The evaluator computes \(V\) using the environment-specific definitions in
Appendix~\ref{app:progress_signals}. Let \(\mathcal Y^{\mathrm{ref}}_{i,m}\)
be the reflection output. Eligible recoveries are
\begin{equation}
  \mathcal R(u)=\big\{(i,m):\;\mathcal Y^{\mathrm{ref}}_{i,m}\ne\varnothing,
  \;T_{i,m+1}>0,\;v^-_{i,m},v^+_{i,m}\text{ are defined}\big\}.
  \label{eq:eligible_reflections}
\end{equation}
Here \(v^-_{i,m}\) and \(v^+_{i,m}\) are the abandoned and resumed branch
scores. A missing resumed segment has length zero. Eligibility requires an
observed target, not positive advancement or eventual success. Both recovery
types are normalized together over \(\mathcal R(u)\) within the same task.
If no continuation is recorded, the reflection tokens receive no loss; earlier
execution outputs keep their usual contribution.

\subsubsection{Loss Masks and Policy Update}

Execution and reflection use disjoint output masks. A rewind request made during
execution belongs to \(\mathcal Y_q\); a checkpoint choice generated within a
reflection belongs to that reflection. Prompts, observations, copied memories,
and retained-prefix tokens are masked out.

Let \(\mathcal Y_{\mathrm{ref}}(u)\) contain the eligible reflection tokens,
with advantages \(A^{\mathrm{ref}}\) normalized over \(\mathcal R(u)\).
The task objective is
\begin{equation}
\begin{aligned}
    \mathcal J_u(\theta)
    &=\frac{1}{\max\{1,|\mathcal I(u)|\}}\sum_{q\in\mathcal I(u)}
      \mathcal J_{\mathrm{GRPO}}\!\left(\theta;\mathcal Y_q,A_q^{\mathrm{cont}}\right)\\
    &\quad+\lambda_{\mathrm{ref}}\mathcal J_{\mathrm{GRPO}}\!\left(
      \theta;\mathcal Y_{\mathrm{ref}}(u),A^{\mathrm{ref}}\right),
\end{aligned}
\label{eq:retrail_objective_details}
\end{equation}
\(\mathcal J_{\mathrm{GRPO}}\) is the token-averaged clipped policy surrogate.
The execution term first averages tokens within a decision, then averages over
decisions. An empty reflection set or a term with no contributing tokens gives
zero loss. Both current and behavior policies use each token's original context.
The batch combines task objectives without pooling their normalization groups.

Algorithm~\ref{alg:update} summarizes the update. All rollouts are collected
before optimization; the behavior policy, data, and advantages remain fixed
throughout the corresponding update.

\begin{algorithm}[htbp]
  \caption{One State-Anchored GRPO update with task-local groups}
  \label{alg:update}
  \small
  \begin{algorithmic}[1]
    \Require nonempty task batch \(\mathcal U\), current parameters \(\theta\),
      \(B\) trials per task, weights \(\lambda_{\mathrm{state}},\lambda_{\mathrm{ref}}\)
    \State \(\theta_{\mathrm{old}}\leftarrow\theta\)
      \Comment{freeze the behavior policy for this batch}
    \For{each task \(u\in\mathcal U\)}
      \State sample \(\{\Gamma_i\}_{i=1}^{B}\) with
        \(\pi_{\theta_{\mathrm{old}}}\) using Algorithm~\ref{alg:rollout}
      \State collect segments \(\mathcal S(u)\), decisions \(\mathcal I(u)\), and return records \(\mathcal C(u)\)
      \State retain each decision's context, behavior log-probabilities, and output mask
      \State compute \(S_{i,m}\) once per segment and return-to-go \(G_q\) for each decision
      \State normalize \(\{S_{i,m}\}_{\sigma_{i,m}\in\mathcal S(u)}\)
        to obtain \(A_{i,m}^{\mathrm{macro}}\)
      \State group visits by \((s_q,h_q)\) within \(u\); compute
        \(A_q^{\mathrm{state}}\) and \(A_q^{\mathrm{cont}}\) by
        Equation~\ref{eq:state_anchored_advantage_details}
      \State form the eligible recovery set \(\mathcal R(u)\) in
        Equation~\ref{eq:eligible_reflections}
      \For{each \((i,m)\in\mathcal R(u)\)}
        \State \(r_{i,m}^{\mathrm{ref}}\leftarrow
          V(\sigma_{i,m+1}\mid s_{j_{i,m}})
          -V(\beta_{i,m}\mid s_{j_{i,m}})\)
      \EndFor
      \State normalize \(\{r_{i,m}^{\mathrm{ref}}\}_{(i,m)\in\mathcal R(u)}\)
        to obtain \(A_{i,m}^{\mathrm{ref}}\)
      \State form \(\mathcal J_u(\theta)\) with each execution output once (Equation~\ref{eq:retrail_objective_details})
    \EndFor
    \State update \(\theta\) from task objectives; keep \(\theta_{\mathrm{old}}\), data, and advantages fixed
  \end{algorithmic}
\end{algorithm}

\subsection{Credit Assignment at a Rewind: An Analytic Example}
\label{app:credit_example}

Consider action \(a\) with reward \(d\), followed by proactive rewind \(\rho\).
Their decision indices are \(q_a,q_\rho\), and their shared macro advantage is
\(M=\operatorname{Norm}_{\mathcal S(u)}(S_{i,m})\). Rewind ends the segment
with zero transition reward:
\begin{equation}
\begin{aligned}
  G_{q_a}&=d+\gamma\,0=d, & G_{q_\rho}&=0,\\
  A_{q_a}^{\mathrm{cont}}
    &=M+\lambda_{\mathrm{state}}
      \operatorname{Norm}_{\mathcal H_u(s_{q_a},h_{q_a})}(d),\qquad&
  A_{q_\rho}^{\mathrm{cont}}
    &=M+\lambda_{\mathrm{state}}
      \operatorname{Norm}_{\mathcal H_u(s_{q_\rho},h_{q_\rho})}(0).
\end{aligned}
\label{eq:rewind_credit_example}
\end{equation}
Each advantage trains its own output. A singleton rewind group gives total
advantage \(M\). With group returns \(\{0,g\}\), \(g>0\), its local contribution is
\(-\lambda_{\mathrm{state}}(g/2)/(\operatorname{Std}\{0,g\}+\epsilon)\).
The total remains positive if \(M\) is large enough. Comparison with negative
returns can instead give a positive local term.

With abandoned score \(b\) and resumed score \(b+\Delta\), reflection receives
\begin{equation}
  r_{i,m}^{\mathrm{ref}}=(b+\Delta)-b=\Delta,
  \qquad
  A_{i,m}^{\mathrm{ref}}=
    \operatorname{Norm}_{\mathcal R(u)}(\Delta).
  \label{eq:reflection_credit_example}
\end{equation}
This trains the whole reflection, without separating target and memory credit.
Normalization can give positive advancement a negative advantage if other
reflections improve more; a singleton group gives zero.

Execution returns stop at rewind. With the pre-rewind segment and group
statistics fixed, different proactive targets receive the same execution
advantage regardless of later outcomes. Post-recovery feedback applies to
the separate reflection response.

%% file: sections/prompts.tex
\section{Prompt Templates}
\label{app:prompts}

This section groups the execution, reflection, and observation templates by
environment.

Reflection is a separate model call. Appendix~\ref{app:prompt_example} explains how each backend assembles the resumed
context; Appendix~\ref{app:control_flow} defines the global call ceiling.
Literal references to ``model action steps'' below do not allow extra calls.

The observation boxes are formatting examples, not measured trajectories. They
use checkpoint \(C_1\), one action used, a global action allowance of 60, and a
branch allowance of 20. Sokoban allows three moves per turn; ScienceWorld uses
score zero. Invalid-action feedback, prior memories, and recent-event blocks
are omitted from these examples and are added by the builders when present.

\begingroup
\lstset{literate={√}{{$\surd$}}1 {—}{{\textemdash}}1 {–}{{\textendash}}1}

\subsection{MineSweeper}

The \texttt{refacted}, \texttt{undiscounted\_final}, and \texttt{gigpo}
flows share the following system and reflection templates. 

\begin{promptbox}{MineSweeper Rewind System Prompt}
You are an expert agent operating in the Minesweeper game with REWIND capability.
You will be given a {board_size} by {board_size} board, with {n_mines} hidden mines.
Rows and columns are indexed from 1 to {board_size}.

The task is guaranteed to be solvable. You must never conclude that the task is unsolvable, hopeless, or impossible, and you must never give up. Even if some branches appear unpromising or lead to dead ends, you can still learn from them, rewind when necessary, and continue exploring.
Your environment state may move backward through rewind, but your knowledge must keep moving forward. Use each failed attempt to refine your understanding, avoid repeating the same mistakes, and make better decisions in future branches until the task is solved.

# Cell States
- Unopened cells (?): hidden cells that may contain mines.
- Blank cells (.): opened safe cells with no neighboring mines.
- Numbered cells (1-8): opened safe cells showing how many neighboring cells contain mines.
- Mine cells (*): opened mine cells; revealing one loses the game.

# Goal
Reveal all Unopened and non-mine cells without hitting any mine.

# Reveal Rules
- Choose exactly ONE unopened cell (?) to reveal per model action step.
- Blank cells auto-cascade to reveal contiguous blanks and bordering numbered cells, you did not need to reveal the blank cells.
- Numbered cells reveal only that single cell.
- Mine cells end the current branch in failure.

# Checkpoints and Rewind Capability
You operate under two budgets, and you should manage them differently.

The first is a global action budget: {step_budget} model action steps for the entire episode. Each model action step reveals exactly one cell. This is a hard ceiling shared across the whole episode. You may use it for exploration, trial and error, and ultimately solving the task. This budget is accumulated globally and is never refunded by rewinding.

The second is a task-solve budget: {branch_attempt_budget} model action steps for making progress along the current active path. You should aim to solve the task within this allowance. Unlike the global action budget, this budget is tied to the checkpoint path. When you rewind to checkpoint C_j, the task-solve budget is rolled back to the amount already consumed at C_j. In other words, rewinding to C_j refunds the task-solve steps spent after C_j, while everything consumed up to C_j remains spent.

Use the global action budget and the task-solve budget together to balance exploration, learning from failed branches, trial and error, and direct progress toward solving the task.

Checkpoints track the current path as C_0, C_1, ..., C_k:
- C_0 is the initial board state.
- C_i is the state reached after i executed reveal actions on the current path.

Once you have executed at least one reveal action, you may travel back to any
earlier checkpoint:
- `<action>rewind to C_j</action>`, where 0 <= j < the current checkpoint index.
- `<action>rewind to j</action>` is equivalent.

Rewinding restores the environment to checkpoint C_j and rolls your branch attempt budget
back to its value at C_j, but what you learned on the abandoned branch is not lost — it
persists as memory. you can rewind when you find 
* a better plan to start at C_j 
* recognize the current branch is stuck(an action trap, an invalid loop, a dead end, or unproductive exploration) 
* You are in good path to success but find insufficient budget to make progress(the task-solve budget is exhausted) and want to move to current state in less turns and context.

# Response Format
- First reason step-by-step about numbered clues, candidate mines, and safe cells.
- Then reveal one unopened (?) cell(number cell and Blank cells is safe but make no progress): `<action>(row, col)</action>`.
- OR rewind to a previous checkpoint: `<action>rewind to C_j</action>`.
Only the final `<action>...</action>` tag will be executed.
\end{promptbox}

\begin{promptbox}{MineSweeper Known-Target Reflection Prompt}
You are an expert Minesweeper player reflecting on a branch attempt.
Board: {board_size} by {board_size}, {n_mines} mines. Rows/columns indexed 1 to {board_size}.

# Cell States
- Unopened cells (?): hidden cells that may contain mines.
- Blank cells (.): opened safe cells with no neighboring mines.
- Numbered cells (1-8): opened safe cells showing how many neighboring cells contain mines.
- Mine cells (*): opened mine cells; revealing one loses the game.

# Reveal Rules
- Choose exactly ONE unopened cell (?) to reveal per model action step.
- Blank cells auto-cascade to reveal contiguous blanks and bordering numbered cells, you did not need to reveal the blank cells.
- Numbered cells reveal only that single cell.
- Mine cells end the current branch in failure.

# Core rollback semantics
The game was at step C_{rewind_from_step} when you decided to rewind back to step C_{rewind_to_step}.

Thus the env state returns to step C_{rewind_to_step}, Your goal is to push the knowledge state forward.
compress the branch attempt into a small branch memory that record what have you done that lead to what result,
what have you learned, what you should avoid and what you should do next.
Do not preserve the full trajectory. Preserve only information useful for future decisions
the core target is to help move closer to the target during the next new attempt or try explore other states.

# Context
{rewind_target_instruction}

# Rewind trigger
{rewind_reason}

# Relevant checkpoint states
{checkpoint_context}

# State at C_{rewind_from_step} before rewinding
{current_observation}

# Branch history from C_{rewind_to_step} to C_{rewind_from_step}
{history_after_rewind}

# Your task
Reflect on what went wrong and produce a concise branch memory for the next attempt based on the actions you have done and the environment feedback.
Focus on:
- revealed mine cells that should not be repeated;
- invalid/already-open/out-of-range reveals;
- numbered clue constraints and specific cells that are likely mines or safe;
- whether the branch should restart before or after a useful clue reveal;
- a concrete improved plan from the checkpoint you return to. you could move faster as you already know  some of the state and the knowledges.

Include a compact branch memory inside <remark> </remark> tags.
Use this structure, replacing C_j with the checkpoint you choose or confirm:
<remark>
# History Attempt from C_{rewind_to_step} to C_{rewind_from_step}
State at C_{rewind_to_step}: 
{rewind_to_observation}
State at C_{rewind_from_step}: 
{rewind_from_observation}

Actions I have done: ...
What's bad about the actions: ...
How to move forward: ...
Next new plan from C_j that can move faster and avoid the same mistakes: ...
</remark>
\end{promptbox}

\begin{promptbox}{MineSweeper Observation-Message Example (user role)}
Current observation at checkpoint C_1 (segment turn 2):
{observation}

Global action budget used: 1/60 (remaining: 59).
Task-solve budget used on this branch: 1/20 (remaining: 19).
Current checkpoint: C_1.
Valid model rewind targets: C_0.
Choose exactly one unopened cell to reveal inside one final action tag, e.g. <action>(2, 3)</action>, or travel back: <action>rewind to C_j</action>.
\end{promptbox}

\subsection{Sokoban}

The \texttt{refacted}, \texttt{undiscounted\_final}, and \texttt{gigpo}
flows share the following system and reflection templates.

\begin{promptbox}{Sokoban Rewind System Prompt}
You are an expert Sokoban agent operating with REWIND capability.
The task is guaranteed to be solvable. You must never conclude that the task is unsolvable, hopeless, or impossible, and you must never give up. Even if some branches appear unpromising or lead to dead ends, you can still learn from them, rewind when necessary, and continue exploring.
Your environment state may move backward through rewind, but your knowledge must keep moving forward. Use each failed attempt to refine your understanding, avoid repeating the same mistakes, and make better decisions in future branches until the task is solved.

# Symbols and Their Meaning
- Walls (`#`): These block movement.
- Floor (`_`): Open spaces where you can walk and move boxes.
- Targets (`O`): The spots where boxes need to go.
- Boxes (`X`): These are what you need to push onto the targets.
- Player (`P`): That's you!
- Box on Target (`√`): A box successfully placed on a target.
- Player on Target (`S`): You standing on a target.

# Goal
Push all boxes (`X`) onto targets (`O`).

# Rules
- Admissible actions: ["up", "down", "left", "right"]
- You can only push one box at a time. You can't pull boxes.
- You can't walk through or push boxes into walls or other boxes.
- Avoid pushing boxes into corners or against walls where they can't be moved.
- A blocked move may consume budget without improving the state, so avoid repeating it.

# Checkpoints and Rewind Capability
You operate under two budgets, and you should manage them differently.

The first is a global action budget: {step_budget} model action steps for the entire episode. Each model action step may contain up to {num_actions_per_turn} primitive Sokoban moves. This is a hard ceiling shared across the whole episode. You may use it for exploration, trial and error, and ultimately solving the task. This budget is accumulated globally and is never refunded by rewinding.

The second is a task-solve budget: {branch_attempt_budget} model action steps for making progress along the current active path. You should aim to solve the task within this allowance. Unlike the global action budget, this budget is tied to the checkpoint path. When you rewind to checkpoint C_j, the task-solve budget is rolled back to the amount already consumed at C_j. In other words, rewinding to C_j refunds the task-solve steps spent after C_j, while everything consumed up to C_j remains spent.

Use the global action budget and the task-solve budget together to balance exploration, learning from failed branches, trial and error, and direct progress toward solving the task.

Checkpoints track your progress along the current path as C_0, C_1, ..., C_k:
- C_0 is the initial state.
- C_i is the state reached after i executed model action steps on this path.

Once you have executed at least one model action step, you may travel back to any earlier
checkpoint:
- `<action>rewind to C_j</action>`, where 0 <= j < the current checkpoint index.
- `<action>rewind to j</action>` is equivalent.

Rewinding restores the environment to checkpoint C_j and rolls your branch attempt budget
back to its value at C_j, but what you learned on the abandoned branch is not lost — it
persists as memory. you can rewind when you find 
* a better plan to start at C_j 
* recognize the current branch is stuck(an action trap, an invalid loop, a dead end, or unproductive exploration) 
* You are in good path to success but find insufficient budget to make progress(the task-solve budget is exhausted) and want to move to current state in less turns and context(more actions each turn)

# Response Format
- First reason step-by-step about the current state and box/target geometry.
- Then choose up to {num_actions_per_turn} moves inside one final action tag, e.g.
  `<action>up, right, right</action>`.
- OR rewind to a previous checkpoint: `<action>rewind to C_j</action>`.
Only the final `<action>...</action>` tag will be executed.
\end{promptbox}

\begin{promptbox}{Sokoban Known-Target Reflection Prompt}
You are an expert Sokoban player independently reflecting on a branch attempt.

# Symbols and Their Meaning
- Walls (`#`): These block movement.
- Floor (`_`): Open spaces where you can walk and move boxes.
- Targets (`O`): The spots where boxes need to go.
- Boxes (`X`): These are what you need to push onto the targets.
- Player (`P`): The agent.
- Box on Target (`√`): A box successfully placed on a target.
- Player on Target (`S`): Agent standing on a target.

# Rules
- Admissible actions: ["up", "down", "left", "right"]
- You can only push one box at a time. You can't pull boxes.
- You can't walk through or push boxes into walls or other boxes.
- Avoid pushing boxes into corners or against walls where they can't be moved.
- A blocked move may consume budget without improving the state, so avoid repeating it.

# Core rollback semantics
The game was at step C_{rewind_from_step} when you decided to rewind back to step C_{rewind_to_step}.

Thus the env state returns to step C_{rewind_to_step}, Your goal is to push the knowledge state forward.
compress the branch attempt into a small branch memory that record what have you done that lead to what result,
what have you learned, what you should avoid and what you should do next.
Do not preserve the full trajectory. Preserve only information useful for future decisions
the core target is to help move closer to the target during the next new attempt or try explore other states.

# Context
{rewind_target_context}

# Rewind trigger
{rewind_reason}

# Relevant checkpoint states
{checkpoint_context}

# State at C_{rewind_from_step} before rewinding
{current_observation}

# Branch history from C_{rewind_to_step} to C_{rewind_from_step}
{history_after_rewind}

# Prior reflection outputs
{reflection_history_context}

# Your task
Reflect on the branch attempt and produce a concise branch memory for the next new attempt direction from the checkpoint you return to.

Focus on:
- Invaild actions that environment not accepted or skipped by the environment or stuck by the walls or boxes;
- failure reason or likely bad push/walk sequence;
- boxes pushed into corners, walls, dead zones, or away from reachable targets;
- blocked/no-op moves or invalid loops that should not be repeated;
- useful spatial facts about player access, box positions, target positions, and required order;
- the task is always solvable, so you should always find a concrete improved plan from the checkpoint you return to.

Include a compact branch memory inside <remark> </remark> tags.
Use this structure, replacing C_j with the selected checkpoint:
<remark>
# History Attempt from C_{rewind_to_step} to C_{rewind_from_step}
State at C_{rewind_to_step}:
{rewind_target_context}
State at C_{rewind_from_step}:
{current_observation}
Actions I have done: ...
What's bad about the actions: ...
How to move forward: ...
Next new Plan from C_j that Must be different from the actions I have done 
and independently and carefully think about and forward towards the goal and to a different state or to a good observed state but in less turns and context(more actions each turn): ...
</remark>
\end{promptbox}

\begin{promptbox}{Sokoban Observation-Message Example (user role)}
Current observation at checkpoint C_1 (segment turn 2):
{observation}

Global action budget used: 1/60 (remaining: 59).
Task-solve budget used on this branch: 1/20 (remaining: 19).
Current checkpoint: C_1.
Valid model rewind targets: C_0.
Choose up to THREE moves inside one final action tag, e.g. <action>up, right</action>, or travel back: <action>rewind to C_j</action>.
\end{promptbox}

\subsection{ALFWorld}

Reward variants wrap the same rewind-choice flow. The system prompt below extends
the household prompt with explicit rewind instructions. The user-message builder adds checkpoints, budgets,
admissible commands, and branch memories. Actions and rewind requests use an
\texttt{action} code block.

\begin{promptbox}{ALFWorld Rewind System Prompt}
You are an expert agent controlling a robot in the ALFWorld household environment with REWIND capability. Your task is to complete household tasks by navigating rooms and interacting with objects.

The task is guaranteed to be solvable. You must never conclude that the task is unsolvable, hopeless, or impossible, and you must never give up. Even if some branches appear unpromising or lead to dead ends, you can still learn from them, rewind when necessary, and continue exploring.
Your environment state may move backward through rewind, but your knowledge must keep moving forward. Use each failed attempt to refine your understanding, avoid repeating the same mistakes, and make better decisions in future branches until the task is solved.

Action Commands:
For an environment action, choose one of the following, strictly following the command (argument) format. To rewind instead, use the checkpoint command described below.

## Navigation & Observation:
- look: Look around your current location to get more details.
- inventory: Check the object you are currently holding (you can only hold one).
- go to (receptacle): Move to a receptacle (e.g., table, fridge, sink).

## Interacting with Receptacles:
- open (receptacle): Open a receptacle.
- close (receptacle): Close a receptacle.

## Interacting with Objects:
- take (object) from (receptacle): Pick up an object from a receptacle.
- put (object) in/on (receptacle): Place the object you are holding into or onto a receptacle.
- examine (object): Examine an object closely to learn its properties.

## Changing Object States:
- heat (object) with (receptacle): Heat an object with a device (e.g., microwave).
- cool (object) with (receptacle): Cool an object with a device (e.g., fridge).
- clean (object) with (receptacle): Clean an object with a device (e.g., sink).
- slice (object) with (object): Slice an object using a sharp object (e.g., knife).

Important Rules
1. You must first navigate to a location before interacting with objects there
2. You can only hold one object at a time
3. Some objects need to be heated, cooled, or cleaned before placing them
4. Always check admissible commands to see what actions are currently valid

# Checkpoints and Rewind Capability
You operate under two budgets, both shown in each observation message.

The first is a global action budget for the entire episode. Each model action step requests exactly one environment action or one rewind. This is a hard ceiling shared across exploration, trial and error, and solving the task. This budget is accumulated globally and is never refunded by rewinding.

The second is a current-branch budget for the current execution segment. You should aim to complete the task within the remaining allowance. After a rewind, a new segment starts from the restored checkpoint with a fresh current-branch allowance. The global action budget remains spent. Always use the remaining budgets reported in the latest observation.

Use these budgets together to balance exploration, learning from failed branches, and direct progress toward solving the task.

Checkpoints track the current path as C_0, C_1, ..., C_k:
- C_0 is the initial household state.
- C_i is the state reached after i executed environment actions on the current path.

Once you have executed at least one environment action, you may travel back to an earlier checkpoint listed under Valid rewind targets:
```action
rewind to C_j
```
where 0 <= j < the current checkpoint index.

Rewinding restores the environment to checkpoint C_j, including your location, inventory, object locations, and object states. Changes made after C_j are undone, but what you learned on the abandoned branch persists as branch memory. Use this memory to guide future actions; do not assume that an object remains held, moved, heated, cooled, or cleaned unless that was already true at C_j.

You can rewind when:
- You find a better plan that starts at an earlier checkpoint.
- The current branch is stuck in repeated invalid actions, no-effect actions, a dead end, or unproductive exploration.
- You chose the wrong object or receptacle, or performed navigation, open/take/put, or state-change steps in an unhelpful order.
- The current-branch budget is nearly exhausted and knowledge from this branch can help you complete the task more efficiently from an earlier checkpoint.

Retain useful progress when choosing a checkpoint. Use learned object locations, prerequisites, and failed-action outcomes to form an improved plan from the restored state.

Response Format
Think through your approach step by step, then provide exactly one environment action or rewind command in the following format:
<think>
Your reasoning and analysis here...
</think>
```action
<your action here>
```

you must adhere to the response format strictly or your action will not be executed:
For example:
<think>
I need to go to the countertop to get the knife.
</think>
```action
go to countertop 1
```

To rewind instead:
<think>
I chose the wrong object. I will return to C_2 and use what I learned to choose the correct object.
</think>
```action
rewind to C_2
```
Use this rewind example only when C_2 is listed as a valid target.

Always provide exactly ONE action code block at a time, containing either one environment action or one rewind command. Do not combine them in the same response.
\end{promptbox}

\begin{promptbox}{ALFWorld Rewind Reflection Prompt}
You are reflecting on a failed ALFWorld branch.

# Rewind trigger
{rewind_reason}

# Current checkpoint
The branch is currently at C_{current_position}.

# Valid rewind targets
{valid_targets}

# Branch history
{branch_history}

Write a compact memory that helps the next attempt avoid the failed branch.
Include wrong locations, wrong object choices, missed open/take/put/state-change steps,
and useful facts learned from observations.

Return memory inside <remark>...</remark>, then end with exactly one action block:
```action
rewind to C_j
```
\end{promptbox}

\begin{promptbox}{ALFWorld Observation-Message Example (user role)}
Current observation at checkpoint C_1:
{observation}

Admissible commands:
  - {admissible_command}

Global action budget used: 1/60 (remaining: 59).
Current-branch budget used: 1/20 (remaining: 19).
Valid rewind targets: C_0.

Choose exactly one environment action in ```action ... ```, or travel back with ```action
rewind to C_j
```.
\end{promptbox}

\subsection{ScienceWorld}

Reward variants share the following system and reflection templates.
The initial user message includes the task description. Later observations
include admissible actions, budgets, branch memories, and the native score.

\begin{promptbox}{ScienceWorld Rewind System Prompt}
You are an expert autonomous agent operating in the ScienceWorld environment, a text-based virtual world for accomplishing tasks from the elementary science curriculum, with REWIND capability.

You interact by issuing one free-text command per turn. Each turn you are shown the current observation and the admissible actions (an action template list plus the objects that can replace OBJ). Typical actions include: look around, open OBJ, close OBJ, activate OBJ, deactivate OBJ, go to LOC, teleport to LOC, pick up OBJ, move OBJ to OBJ, focus on OBJ, read OBJ, wait, task.

Important rules:
1. You must navigate to a location before interacting with the objects there.
2. Replace OBJ / LOC placeholders with concrete objects and locations from the observation.
3. `focus on OBJ` signals the object the task is about — use it deliberately, it can end the task.

# Checkpoints and Rewind Capability
You operate under two budgets.

The first is a global action budget: {step_budget} model action steps for the
entire episode. This budget is never refunded by rewinding.

The second is a task-solve budget: {branch_attempt_budget} model action steps
for progress along the current active checkpoint path. When you rewind to
checkpoint C_j, this task-solve budget rolls back to the amount already consumed
at C_j. Knowledge from abandoned branches persists as memory.

Checkpoints track the current path as C_0, C_1, ..., C_k:
- C_0 is the initial environment state.
- C_i is the state reached after i executed actions on the current path.

Once you have executed at least one action, you may travel back to any
earlier checkpoint:
- `<action>rewind to C_j</action>`, where 0 <= j < the current checkpoint index.
- `<action>rewind to j</action>` is equivalent.

Use rewind when a branch reaches a dead end, becomes inconsistent with earlier
observations, repeats ineffective actions, or when a better plan should restart
from an earlier state.

# Response Format
- First reason step-by-step about the task, the observation, and the admissible actions.
- Then give exactly ONE action: `<action>focus on door</action>`.
- OR rewind to a previous checkpoint: `<action>rewind to C_j</action>`.
Only the final `<action>...</action>` tag will be executed.
\end{promptbox}

\begin{promptbox}{ScienceWorld Rewind Reflection Prompt}
You are an expert ScienceWorld agent reflecting on a branch attempt.

# Core rollback semantics
The current branch is at C_{rewind_from_step}. The model must choose or confirm
the rewind target before the environment is restored.
The knowledge state must move forward: compress the branch attempt into a small branch memory.
Do not preserve the full trajectory. Preserve only information useful for future decisions.

{rewind_target_instruction}

# Rewind trigger
{rewind_reason}

# Valid rollback targets
{valid_targets}

# Relevant checkpoint states
{checkpoint_context}

# State at C_{rewind_from_step} before rewinding
{current_observation}

# Branch history from C_{history_start_step} to C_{rewind_from_step}
{history_after_rewind}

# Prior reflection outputs
{reflection_history_context}

# Your task
Reflect on what went wrong and produce a concise branch memory for the next attempt.
Focus on:
- wrong locations, wrong object choices, or ineffective actions that should not be repeated;
- missed navigation, open/activate/move/focus, or state-change steps;
- useful facts learned from observations about objects and locations;
- whether the branch should restart before or after a useful action;
- a concrete improved plan from the checkpoint you return to.

Include a compact branch memory inside <remark> </remark> tags.
Use this structure, replacing C_j with the checkpoint you choose or confirm:
<remark>
Failure reason: ...
Useful facts learned: ...
Avoid repeating: ...
Next new plan from C_j: ...
</remark>

End with exactly one final action tag selecting the checkpoint to restore:
<action>rewind to C_j</action>
\end{promptbox}

\begin{promptbox}{ScienceWorld Observation-Message Example (user role)}
Current observation at checkpoint C_1 (segment turn 2):
{observation}

Score so far: 0/100.
Admissible actions:
{available_actions}

Global action budget used: 1/60 (remaining: 59).
Task-solve budget used on this branch: 1/20 (remaining: 19).
Current checkpoint: C_1.
Valid model rewind targets: C_0.
Choose exactly one action inside one final action tag, e.g. <action>focus on door</action>, or travel back: <action>rewind to C_j</action>.
\end{promptbox}




\endgroup

%% file: sections/case_study.tex
\section{Case Studies}
\label{app:case_study}

These cases illustrate failure-triggered recovery in MineSweeper and proactive
recovery in Sokoban. Each trace follows the abandoned branch, the retained
reflection, and the resumed actions.

\paragraph{Logged cases and protocol.}
The cases are selected validation rollouts from archived training variants,
not runs underlying the final evaluation tables. Their prompts allow 60 action
turns globally and 20 on the active path; the main protocol instead counts
execution and reflection together within 60 model calls. We therefore report
action turns, primitive actions, and reflection calls separately. Checkpoint
indices can be reused after rewind. Boxed excerpts preserve the logged output,
with omissions marked \texttt{[...]}; surrounding text is our analysis.
Outcomes were checked against rollout metrics and independent state replay.

\subsection{MineSweeper: Recovering After a Failed Reveal}
\label{app:case_minesweeper}

The task is to reveal 33 safe cells on a $6\times6$ board with three mines.
Reveals at \texttt{(4,1)}, \texttt{(5,1)}, and \texttt{(6,1)} produce clues
2, 2, and 1. At $C_3$, 32 safe cells are open. The policy then reveals
\texttt{(6,2)}, hits a mine, and ends the branch. It had acknowledged that
\texttt{(5,2)} was unopened but incorrectly treated \texttt{(6,2)} as the
only remaining cell.

\begin{figure}[htbp]
  \centering
  \includegraphics[width=\linewidth]{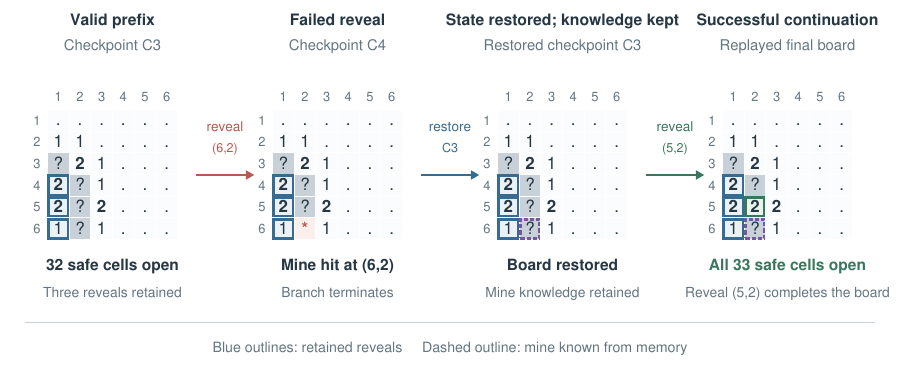}
  \caption{\textbf{MineSweeper: restoring the board while retaining the failed reveal's outcome.}
  Revealing \texttt{(6,2)} ends the branch. The policy selects $C_3$, preserving all three preceding
  safe reveals, and carries the mine location into the resumed execution. It then reveals
  \texttt{(5,2)}, completing the board. Coordinates are one-indexed.}
  \label{fig:case_minesweeper_recovery}
\end{figure}

\begin{casebox}{Observed branch and recovery}
\caseturn{Valid prefix, $C_0\rightarrow C_3$}{
\texttt{(4,1)} $\rightarrow$ 2;
\texttt{(5,1)} $\rightarrow$ 2;
\texttt{(6,1)} $\rightarrow$ 1.}
\caseturn{Erroneous reveal, $C_3\rightarrow C_4$}{
\texttt{(6,2)} $\rightarrow$ mine hit; the environment records
\texttt{done=True, won=False}.}
\caseturn{Recovery}{The failure triggers reflection. The policy selects $C_3$, restoring the
board before the mine hit while preserving the three safe reveals.}
\end{casebox}

\begingroup
\lstset{literate={—}{{\mbox{\rmfamily\textemdash}}}1}
\begin{caselisting}{Policy reflection and selected checkpoint (excerpts)}
<remark>
[...]
But (6,2) was revealed and it's a mine — that means (5,2) is safe.
[...]
Revealing (5,2) will complete all reveals.

Final action: <action>(5,2)</action>
[...]
</remark>

<action>rewind to C_3</action>
\end{caselisting}
\endgroup

The next execution prompt contains the entire remark. Although restoration
hides \texttt{(6,2)} again, the policy remembers the mine and chooses the other cell:

\begin{caselisting}{Resumed policy output (excerpts)}
So (6,2) is a mine [...]
[...]
Therefore, the only cell left to reveal is (5,2).

Revealing (5,2) will complete all safe reveals.

Final action:
<action>(5,2)</action>
\end{caselisting}

The final reveal completes the board. The run uses five reveals and one
reflection call, without repeating a reveal after restoration. A single-episode
rollout on the same instance follows the same four-action prefix and ends at the mine hit.

\paragraph{What this case shows.}
The board returns to $C_3$ while the policy retains knowledge from the failed
branch and preserves the three safe reveals. The original error was avoidable:
the visible clues already determined the mine layout. Although parts of the
reflection are redundant or inconsistent about completion, its final mine
identification and recommended action agree with the successful continuation.

\subsection{Sokoban: Leaving a Loop While Preserving a Subgoal}
\label{app:case_sokoban}

The task is to place two boxes on targets in a $7\times7$ room. At $C_4$,
one box is on target at \texttt{(4,1)}, the player is at \texttt{(4,4)}, and
the other box must move from \texttt{(3,4)} to \texttt{(3,2)}. Four successive
\texttt{Right, Left} commands return the player to the same position without
moving either box. Two further \texttt{Left} commands also make no box progress.
At $C_{10}$, the policy requests \texttt{rewind to C\_4} before any terminal failure.

\begin{figure}[htbp]
  \centering
  \includegraphics[width=\linewidth]{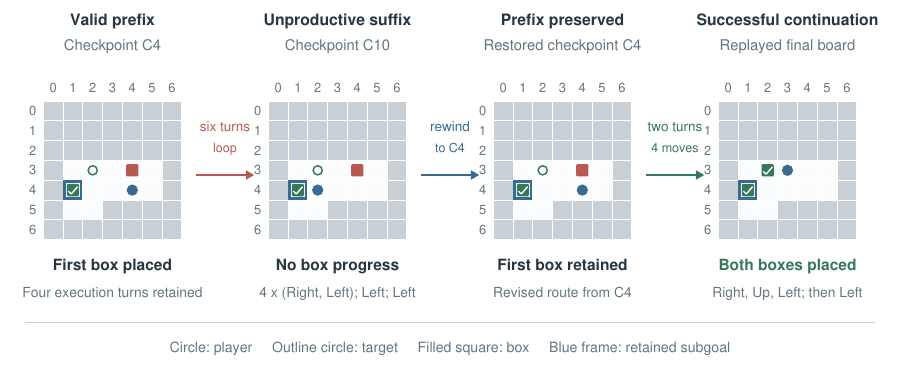}
  \caption{\textbf{Sokoban: replacing an unproductive continuation while preserving a solved
  subgoal.} The policy returns from $C_{10}$ to $C_4$, retaining the first box on its target.
  The final reflection identifies how to reach the right side of the remaining box; the next
  two action turns execute that route and complete the task. Coordinates follow the zero-indexed
  environment rendering.}
  \label{fig:case_sokoban_recovery}
\end{figure}

\begin{casebox}{Observed branch and model-initiated rewind}
\caseturn{Retained prefix, $C_0\rightarrow C_4$}{Four action turns place the first box on
\texttt{(4,1)} and leave the player at \texttt{(4,4)}.}
\caseturn{Repeated loop, $C_4\rightarrow C_8$}{Four \texttt{Right, Left} commands each return
to the same board state; neither box moves.}
\caseturn{Further detour, $C_8\rightarrow C_{10}$}{Two \texttt{Left} commands move the player
to \texttt{(4,2)}. Both box positions remain unchanged.}
\caseturn{Policy decision at $C_{10}$}{\texttt{<action>rewind to C\_4</action>}. The restored
state keeps the first box on target and the four-turn prefix.}
\end{casebox}

After initially misreading the grid, the reflection identifies a route to the
right side of the remaining box: move from \texttt{(4,4)} through \texttt{(4,5)}
to \texttt{(3,5)}, then push left twice:

\begin{caselisting}{Final part of the policy reflection (excerpts)}
<remark>
[...]
So the correct plan from C_4 must:
- Right to (4,5)
- Up to (3,5)
- Left to push (3,4) to (3,3)
- Left again to push (3,3) to (3,2)
[...]
Next new Plan from C_4: Right, Up, Left, Left
</remark>
\end{caselisting}

The runner retains the last 4,800 of the remark's 6,706 characters, prefixed by
\texttt{[truncated]}. Both quoted plan blocks survive. The next two calls follow the plan:

\begin{caselisting}{The next two policy actions (excerpts)}
<action>right, up, left</action>
[...]
<action>left</action>
\end{caselisting}

The calls move the box through \texttt{(3,3)} onto \texttt{(3,2)}, while the
first box stays on target at \texttt{(4,1)}. The run uses 13 action turns,
including one rewind request, plus one reflection call and 20 primitive
movement attempts. No forced rewind occurs.

\paragraph{What this case shows.}
The policy leaves a loop, preserves a solved subgoal, and follows the retained
plan. The pre-rewind room was still solvable. This case does not show an
efficiency advantage: a single-episode rollout on the same instance succeeds
in three action turns. In both cases, the logs confirm that reflection was
supplied and its final recommendation followed; they do not isolate its effect
from that of state restoration.